\documentclass{article}

\usepackage[preprint]{neurips_2026}        %

\usepackage[utf8]{inputenc}
\usepackage[T1]{fontenc}
\usepackage{hyperref}
\usepackage{url}
\usepackage{booktabs}
\usepackage{amsfonts}
\usepackage{amsmath}
\usepackage{amssymb}
\usepackage{nicefrac}
\usepackage{microtype}
\usepackage{xcolor}
\usepackage{graphicx}
\usepackage{subcaption}
\usepackage{multirow}
\usepackage{array}
\usepackage{tabularx}
\newcolumntype{Y}{>{\centering\arraybackslash}X}
\usepackage{makecell}
\usepackage{algorithm}
\usepackage{algpseudocode}
\usepackage{enumitem}
\usepackage[capitalize]{cleveref}
\usepackage[normalem]{ulem}
\usepackage{float}

\graphicspath{{img/}{img/teaser/}{img/pipeline/}{img/results/}{img/ablation/}}

\newcommand{\cset}{\mathcal{C}}                       %
\newcommand{\concept}{c}                              %
\newcommand{\preserveset}{\mathcal{P}}                %
\newcommand{\Nconcepts}{N}                            %

\newcommand{\Tlat}{T_{\mathrm{lat}}}                  %
\newcommand{\dimm}{d}                                 %
\newcommand{\dimT}{d_{\mathrm{T5}}}                   %

\newcommand{\textenc}{\mathcal{E}}                    %
\newcommand{\textemb}{\Phi}                           %
\newcommand{\xattnK}{W^{(\ell)}_{K}}                  %
\newcommand{\xattnV}{W^{(\ell)}_{V}}                  %
\newcommand{\ffnone}{W^{(\ell)}_{\mathrm{ffn}_0}}     %

\newcommand{\loraB}[1]{B_{#1}}
\newcommand{\loraD}[1]{D_{#1}}
\newcommand{\Wlora}{\Delta W}
\newcommand{\phic}{\phi_{\concept}}                   %
\newcommand{\fnorm}{\bar{f}}                          %
\newcommand{\tnorm}{\bar{t}}                          %

\newcommand{\Lesd}{\mathcal{L}_{\mathrm{neg}}}
\newcommand{\Lmax}{\mathcal{L}_{\mathrm{max\text{-}leak}}}
\newcommand{\Lmot}{\mathcal{L}_{\mathrm{motion}}}
\newcommand{\Ltsm}{\mathcal{L}_{\mathrm{tsm}}}
\newcommand{\Ltot}{\mathcal{L}}

\newcommand{\alphaTS}{\alpha_{\mathrm{TPD}}}
\newcommand{\Wuce}{W_{\mathrm{e}}}                    %
\newcommand{\Worig}{W_{0}}
\newcommand{\Wts}{W_{\mathrm{CR}}}
\newcommand{\tsplit}{t^{\star}}

\newcommand{\motionset}{\mathcal{M}}
\newcommand{\anchorMA}{\bar{\mathbf{c}}}

\newcommand{\anchorvec}{\mathbf{h}^{\star}_{\concept}}
\newcommand{\promptemb}{\mathbf{h}_{p}}
\newcommand{\gate}{g_{\concept}}
\newcommand{\routau}{\tau_{\concept}}
\newcommand{\routT}{T_{\concept}}

\newcommand{\meanframeacc}{\mathrm{mean\text{-}frame\text{-}acc}}
\newcommand{\maxframeacc}{\mathrm{max\text{-}frame\text{-}acc}}
\newcommand{\reactgap}{\mathrm{Gap}_{\mathrm{react}}}

\newcommand{\Lpeak}{\Lmax}            %
\newcommand{\Lneut}{\Lmot}            %
\newcommand{\Lflick}{\Ltsm}           %

\newcommand{\peakframeacc}{\maxframeacc}   %

\newcommand{\embMPE}{\anchorMA}       %
\newcommand{\alphaTPD}{\alphaTS}      %
\newcommand{\Wcr}{\Wts}               %
\newcommand{\refvec}{\anchorvec}      %

\definecolor{fadeF}{HTML}{07AEEA}
\definecolor{fadeA}{HTML}{13C6CF}
\definecolor{fadeD}{HTML}{1FDDB3}
\definecolor{fadeE}{HTML}{2BF598}
\definecolor{oursrow}{HTML}{E8F6F1}

\newcommand{\ours}{\rowcolor{oursrow}}

\usepackage{colortbl}
\usepackage{xcolor}
\title{\textcolor{fadeF}{F}\textcolor{fadeA}{A}\textcolor{fadeD}{D}\textcolor{fadeE}{E}: \textcolor{fadeF}{F}rame-\textcolor{fadeA}{A}ware \textcolor{fadeD}{D}iffusion-Transformer-based Multi-Concept \textcolor{fadeE}{E}rasure for Video Unlearning}

\author{%
  \parbox[t]{0.3\textwidth}{\centering
    Yuchen Li\textsuperscript{*} \\
    Peking University \\
    \texttt{yuchenli@stu.pku.edu.cn}
  }
  \And
  \parbox[t]{0.3\textwidth}{\centering
    Kaiyuan Deng\textsuperscript{*} \\
    University of Arizona \\
    \texttt{kaiyuan0415@arizona.edu}
  }
  \AND
  \parbox[t]{0.3\textwidth}{\centering
    Chaoran Feng \\
    Peking University \\
  }
  \And
  \parbox[t]{0.3\textwidth}{\centering
    Zhenyu Tang \\
    Peking University \\
  }
  \And
  \parbox[t]{0.3\textwidth}{\centering
    Li Yuan\textsuperscript{$\dag$} \\
    Peking University \\
    \texttt{yuanli-ece@pku.edu.cn}
  }
}

\begin{document}
\maketitle
{\renewcommand{\thefootnote}{}\footnotetext{$^*$Equal contribution. \quad $^\dagger$Corresponding author.}}

\begin{abstract}
Text-to-video (T2V) diffusion models can reproduce copyrighted, violent, or explicit content, which motivates concept erasure: removing designated concepts from a pretrained model while preserving its behavior on everything else. Existing T2V erasure methods leave two problems open. Their frame-agnostic suppression can leave isolated frames in which an erased concept resurfaces, a \emph{frame-reactivation gap} that clip-level averages obscure; and they are usually evaluated with one target concept or category at a time. We propose Frame-Aware Diffusion Erasure (\textbf{\textcolor{fadeF}{F}\textcolor{fadeA}{A}\textcolor{fadeD}{D}\textcolor{fadeE}{E}}), a multi-concept video unlearning framework. FADE first applies a joint closed-form key/value edit that suppresses all target concepts, then trains per-concept frame-aware low-rank adapters whose strength is gated by the frame index and the denoising timestep to remove residual per-frame leakage. Each adapter is trained with the other targets' prompts as hard negatives, which keeps the concept-specific components of different adapters well separated, and a similarity-based soft router combines the adapters according to the prompt. With 16 concepts (objects, artistic styles, and nudity) erased from a single Wan2.1-T2V-1.3B backbone, FADE reduces the residual accuracy on the object benchmark to 4.9\%, against 15.5\% for the strongest of eight baselines, while keeping the VBench average within 0.9\% of the unedited model. The ranking is unchanged under a VLM judge and a blinded human study, and the advantage over the strongest baseline carries over to prompts that combine several erased concepts, to 30 simultaneously erased celebrity identities, and to Wan2.1-T2V-14B, CogVideoX-2B, and HunyuanVideo-1.5.
\end{abstract}

\begin{figure}[t]
      \centering
      \includegraphics[width=0.87\linewidth]{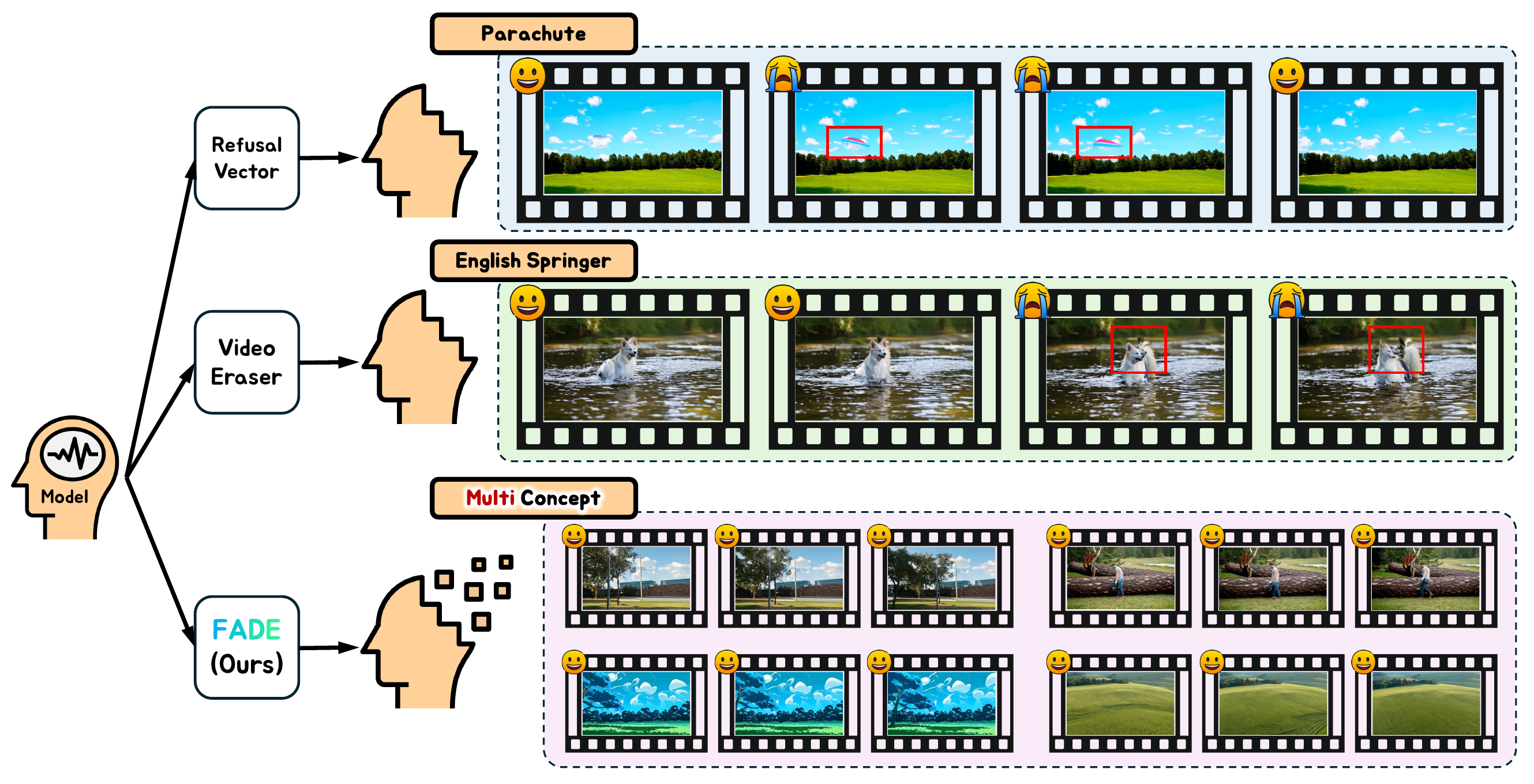}
\caption{\textbf{Frame reactivation and multi-concept erasure.} Prior T2V erasure methods apply the same suppression to every frame, and an erased concept can resurface mid-clip (red boxes; top: Refusal Vector on \emph{parachute}, middle: VideoEraser on \emph{English springer}). FADE erases several concepts from a single backbone, and in these examples they stay suppressed in every frame (bottom).}
      \label{fig:teaser}
\end{figure}
\section{Introduction}
\label{sec:intro}

Text-to-video (T2V) diffusion models~\citep{ho2022video,singer2022make,wang2023modelscope,guo2023animatediff,chen2024videocrafter2,yang2024cogvideox,kong2024hunyuanvideo,zheng2024open,wan2025wan,liu2024sora} turn short natural-language prompts into visually compelling and temporally coherent clips, and open-weight systems such as CogVideoX~\citep{yang2024cogvideox}, HunyuanVideo~\citep{kong2024hunyuanvideo}, Open-Sora~\citep{zheng2024open}, and Wan~\citep{wan2025wan} are now widely deployed. The same capability lets these models reproduce copyrighted material, generate violent content, support video-based disinformation, or synthesize sexually explicit imagery~\citep{schramowski2023safe,brundage2018malicious,solaiman2021process,weidinger2021ethical}. \emph{Machine unlearning}~\citep{cao2015towards,bourtoule2021machine,nguyen2022survey,xu2024machine} addresses these risks by removing a model's knowledge of designated concepts while preserving its generative capacity on everything else.

Concept erasure is well studied for text-to-image (T2I) diffusion, with methods based on fine-tuning~\citep{gandikota2023erasing,lu2024mace}, closed-form weight editing~\citep{gandikota2024unified,deng2026forgetmany}, and inference-time guidance~\citep{schramowski2023safe}. For T2V models the field is younger. SAFREE~\citep{yoon2025safree} and VideoEraser~\citep{xu2025videoeraser} intervene on prompt embeddings and the denoising process without changing weights; \citet{liu2024unlearning} and T2VUnlearning~\citep{ye2025t2vunlearning} fine-tune the model for persistent erasure; and the low-rank refusal vector of \citet{facchiano2025video} and ICE~\citep{biswas2025now} edit weights in closed form.

These methods leave two problems open. \textbf{(i) Frame reactivation.} Their suppression signal is frame-agnostic: the same edit or guidance applies to every frame, and a concept that is removed from most of a clip can resurface in a few frames mid-clip, often with visible flicker (\cref{fig:teaser}). Clip-level metrics average this failure away. Existing methods also show frame-to-frame inconsistency~\citep{xu2025videoeraser}, near-static outputs~\citep{yoon2025safree}, blurring~\citep{facchiano2025video}, or degraded benign generation~\citep{ye2025t2vunlearning}, and T2I erasure methods ported to a video DiT leave most of the target in place (\cref{sec:exp:object}). \textbf{(ii) Multi-concept removal.} Existing T2V methods are evaluated with one target concept or one unsafe category at a time~\citep{xu2025videoeraser,yoon2025safree,facchiano2025video,ye2025t2vunlearning}. Deployment instead requires a single model that forgets many concepts at once, and training one shared adapter on all targets severely degrades generation (\cref{sec:exp:ablation}).

We propose \textbf{\textcolor{fadeF}{F}rame-\textcolor{fadeA}{A}ware \textcolor{fadeD}{D}iffusion \textcolor{fadeE}{E}rasure} (\textbf{\textcolor{fadeF}{F}\textcolor{fadeA}{A}\textcolor{fadeD}{D}\textcolor{fadeE}{E}}), a multi-concept unlearning framework for text-conditioned video diffusion transformers. For \textbf{(i)}, FADE combines a global and a frame-level stage. \emph{Concept Redirect} (CR) is a joint closed-form edit of the key/value projections that consume text conditioning, extended with motion-paraphrased concept embeddings and a timestep-dependent edit strength. CR is frame-agnostic and leaves residual per-frame leakage, which \emph{Frame-Aware Erasure} (FAE) removes with a per-concept low-rank adapter whose frame-aware branch is gated by the latent frame index and the denoising timestep and is trained with a loss on the worst-leaking frames. For \textbf{(ii)}, each adapter is trained to act as the identity on the prompts of every other erased concept; the resulting experts share a common suppression template but have well-separated concept-specific components, and a prompt-level soft router (\emph{C-MoE}) combines experts according to prompt similarity. Experts are trained independently, so training parallelizes across concepts.

Our contributions are:
\begin{itemize}[leftmargin=*,itemsep=1pt,topsep=2pt]
    \item \textbf{Diagnostic.} We formalize the frame-reactivation gap, the difference between the peak-frame and clip-mean concept presence, and show that it exposes failures clip-level accuracy misses: removing FADE's frame conditioning leaves clip-mean accuracy almost unchanged but triples the gap.
    \item \textbf{Method.} We propose FADE, which to our knowledge is the first framework that erases multiple concepts from a single T2V model, combining a joint closed-form edit, frame- and timestep-gated per-concept adapters, and similarity-based composition at inference.
    \item \textbf{Evidence.} With 16 concepts erased from one Wan2.1 backbone, FADE reduces residual object accuracy to $4.9\%$ (strongest of eight baselines: $15.5\%$) with the VBench average within $0.9\%$ of the unedited model, and has the lowest attack success rate under four jailbreak suites. It also leads under VLM and human judges, on compositional prompts, for 30 identities, and on four backbones.
\end{itemize}

\section{Related Work}
\label{related}

\paragraph{Concept erasure for image generation.}
Concept-erasure methods for T2I diffusion fall into five groups. \textbf{Fine-tuning} methods update weights to suppress a target, including ESD~\citep{gandikota2023erasing}, CA~\citep{kumari2023ablating}, FMN~\citep{zhang2024forget}, SalUn~\citep{fan2023salun}, AGE~\citep{bui2025fantastic}, EraseDiff~\citep{wu2025erasing}, Sculpting Memory~\citep{li2025sculpting}, and Fine-Grained Erasure~\citep{thakral2025fine}, and ANT~\citep{li2025ant} steers the denoising trajectory during fine-tuning to sidestep unwanted concepts. \textbf{Closed-form editing} solves for new projection weights directly and scales to many concepts, as in UCE~\citep{gandikota2024unified}, RECE~\citep{gong2024reliable}, and ScaPre~\citep{deng2026forgetmany}. MACE~\citep{lu2024mace}, the closest T2I antecedent of FADE, combines a closed-form cross-attention refinement with per-concept LoRAs that are fused into one model; FADE instead keeps per-concept experts separate, gates them by frame and timestep, and composes them at inference. \textbf{Adapter-based} methods such as SPM~\citep{lyu2024one}, GLoCE~\citep{lee2025localized}, and EraseAnything~\citep{gao2025eraseanything} attach low-rank or gated modules; EraseAnything is the first to target rectified-flow transformers such as Flux. \textbf{Inference-time} methods such as SLD~\citep{schramowski2023safe}, AMG~\citep{chen2024towards}, GIE~\citep{chen2024growth}, and Concept Replacer~\citep{zhang2025concept} steer generation without weight updates. \textbf{Neuron pruning}, including ConceptPrune~\citep{chavhan2024conceptprune}, EFCS~\citep{shirkavand2025efficient}, and FIA~\citep{deng2026forgetitall}, removes concept-specific units. FADE builds on two of these primitives, a UCE-style closed-form edit and ESD-style negative guidance, and we evaluate direct ports of UCE, ESD, and MACE to a video backbone in \cref{sec:exp:object}.

\paragraph{Concept erasure for video generation.}
\emph{Inference-time} methods leave weights unchanged: SAFREE~\citep{yoon2025safree} projects toxic token embeddings away from an unsafe subspace and filters the denoising process, and VideoEraser~\citep{xu2025videoeraser} adjusts selected prompt embeddings and adds adversarially resilient noise guidance. \emph{Fine-tuning} methods change parameters persistently: \citet{liu2024unlearning} transfer text-encoder unlearning to T2V models, and T2VUnlearning~\citep{ye2025t2vunlearning} combines negatively guided velocity prediction with prompt augmentation, mask-based localization, and concept-preservation regularization. \emph{Training-free weight updates} include the low-rank refusal vector of \citet{facchiano2025video}, which extracts a concept direction from paired safe/unsafe prompts by contrastive low-rank factorization and folds it into the weights, and ICE~\citep{biswas2025now}, a one-shot spectral weight edit. All of these apply the same suppression to every frame and are evaluated with one target concept or category at a time; FADE adds frame- and timestep-dependent suppression and multi-concept composition.

\paragraph{Adapter composition.}
Sparsely gated mixture-of-experts layers~\citep{shazeer2017moe} route inputs to a subset of experts, and LoRA modules~\citep{hu2022lora} have been composed with learned gates~\citep{wu2024mole} or gradient-free weights~\citep{huang2024lorahub}. C-MoE instead uses a fixed similarity rule computed once per prompt, and its experts are trained with each other's concepts as negatives so that their sum remains selective.

\section{Method}
\label{sec:method}

\begin{figure}[t]
  \centering
  \includegraphics[width=\textwidth]{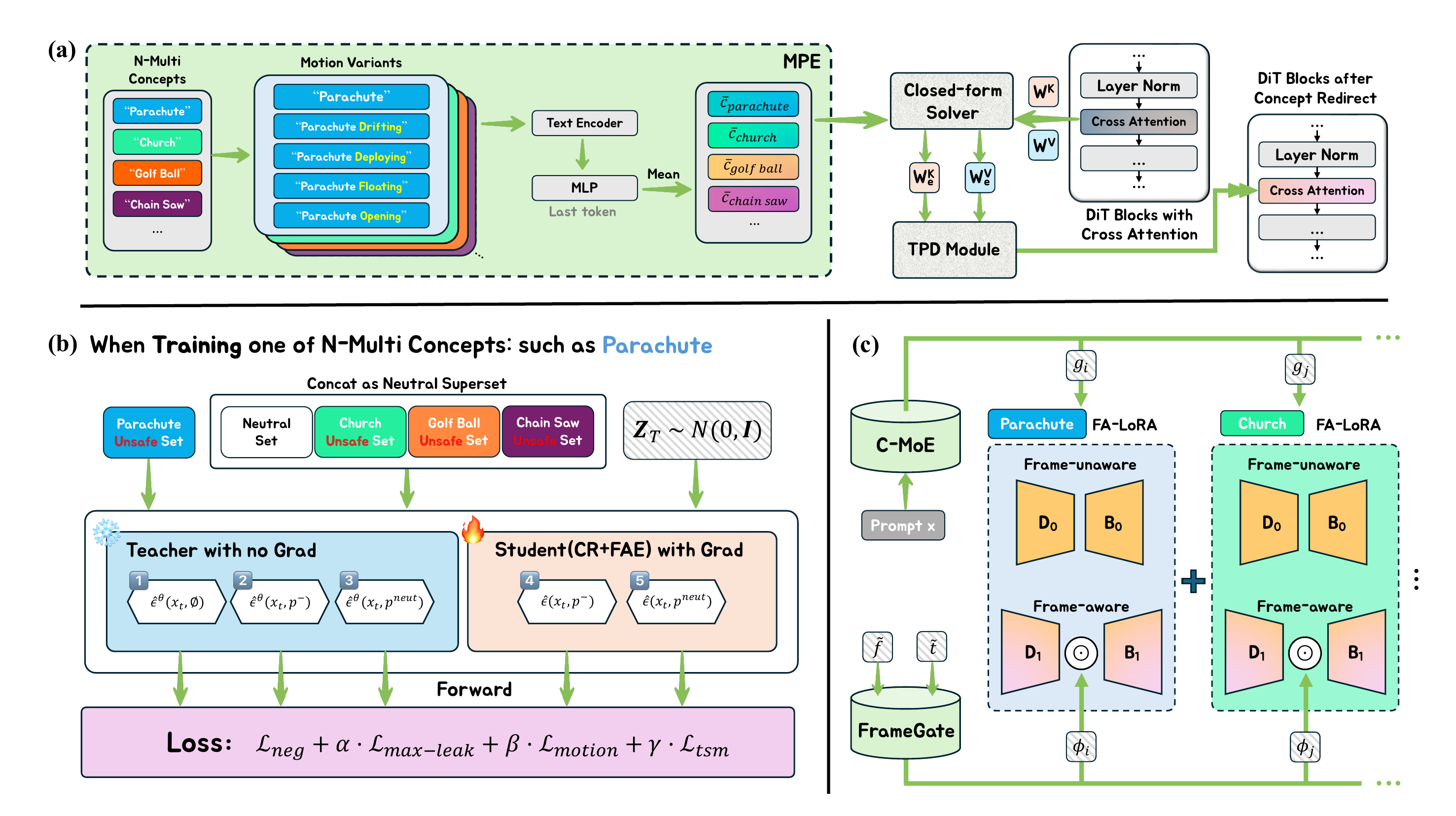}
  \caption{\textbf{FADE overview.} \textbf{(a)} Concept Redirect: motion-paraphrased concept embeddings enter a joint closed-form K/V solve, and TPD blends the edited and original projections over timesteps. \textbf{(b)} FAE training for one concept (here \emph{parachute}): the unsafe prompts of every other erased concept join the neutral set of $\Lneut$ as hard negatives; a frozen CR teacher supplies the targets. \textbf{(c)} At inference, FrameGate modulates the frame-aware branch of each FA-LoRA and the C-MoE router scales each expert by its prompt gate $g_c$.}
  \label{fig:pipeline}
  \vspace{-1.em}
\end{figure}

FADE has three components (\cref{fig:pipeline}). \textbf{Concept Redirect} (CR; \cref{sec:method:cr}) is a joint closed-form edit of the key/value projections that consume text conditioning. It extends the closed-form editor of UCE~\citep{gandikota2024unified} with a \emph{Motion-Paraphrase Embedding} (MPE), which averages a concept's embedding over motion-paraphrased prompts, and a \emph{Texture-Phase Decay} (TPD), which lowers the edit strength during the texture-refinement timesteps. \textbf{Frame-Aware Erasure} (FAE; \cref{sec:method:fae}) attaches a per-concept \emph{frame-aware low-rank adapter} (FA-LoRA) to the FFN path, gated per frame by a small MLP, \emph{FrameGate}, and trained with a four-term temporal objective. \textbf{C-MoE routing} (\cref{sec:method:cmoe}) activates each per-concept expert according to the prompt at inference, so that several concepts can be erased at once.

\subsection{Problem setup and diagnostic}
\label{sec:method:setup}

\paragraph{Setting.}
Let $f_\theta$ be a text-to-video diffusion transformer (DiT)~\citep{peebles2023dit} with frozen text encoder $\textenc$ and text-embedding MLP $\textemb$. FADE edits three per-block linear maps of $f_\theta$: the key/value projections $\xattnK,\xattnV\in\mathbb{R}^{\dimm\times\dimm}$ that map the text context into attention (the cross-attention K/V in Wan2.1), and the first FFN linear map $\ffnone$. Given erase concepts $\cset_e=\{\concept_1,\dots,\concept_{\Nconcepts}\}$ and a preserve set $\preserveset$, we call the task \emph{multi-concept erasure} when a single edited model must (a) forget every $\concept_i$ when it is prompted on its own, (b) forget all targets that co-occur in one prompt, and (c) leave generations on $\preserveset$ and on neutral prompts unchanged. Prior T2V methods address (a) and (c) for $\Nconcepts{=}1$; we evaluate (a) and (c) in \cref{sec:exp:object,sec:exp:style,sec:exp:unsafe} and (b) in \cref{sec:exp:multi}.

\paragraph{Frame Reactivation.}
A per-frame judge yields concept-presence confidence $s_V[f]\in[0,1]$. The clip-level mean $\meanframeacc(V)=(1/F)\sum_f s_V[f]$ used by existing T2V unlearning evaluations can hide a concept that is absent from most frames but present in a few. We complement it with the peak-frame accuracy $\peakframeacc(V) = \max_f s_V[f]$ and the Reactivation Gap
\begin{equation}
\reactgap(V) \;=\; \peakframeacc(V) - \meanframeacc(V),
\label{eq:gap}
\end{equation}
which exposes concepts that resurface in a few frames (reported in \cref{sec:exp:ablation}).

\subsection{Concept Redirect}
\label{sec:method:cr}

CR is a closed-form edit that suppresses all erase concepts jointly in about one minute. On its own it leaves the target concepts largely visible in the generated videos (\cref{fig:appendix-grid-1}, row ``w/o FAE''), but it moves the backbone toward the erasure target, and FAE trained on top of it reaches a lower residual accuracy at a lower fidelity cost than FAE trained on the raw backbone.

\paragraph{Base solver.}
We follow the closed-form K/V editing recipe of UCE~\citep{gandikota2024unified}: for every block $\ell$ and projection $W\!\in\!\{\xattnK,\xattnV\}$ we solve a joint ridge regression that redirects each erase concept $\concept_i$ to a guide anchor $\concept_i^{\star}$ while preserving $W$ on $\preserveset$, taking concept references at the post-MLP last-content-token $\mathbf{c}_i = \textemb(\textenc(\concept_i))_{[-1]}$ (closed form and K/V edit-locus argument in \cref{app:cr-derivation,app:edit-locus}). The solve is joint over all $\Nconcepts$ concepts and is applied once at every block. We add two video-specific changes, one on the input side and one on the timestep side.

\paragraph{Input axis: Motion-Paraphrase Embedding.}
In video, a concept also appears through motion variants (e.g.\ ``parachute drifting'') whose text context lies off the bare-concept direction. We replace $\mathbf{c}_i$ with $\embMPE_i = \tfrac{1}{K+1}\big(\mathbf{c}_i + \sum_{m\in\motionset(\concept_i)}\mathbf{c}_m\big)$, the average of the bare form and $K$ motion-paraphrased variants $\motionset(\concept_i)=\{m_1,\dots,m_K\}$, with $\mathbf{c}_m = \textemb(\textenc(m))$ at the same last-content-token reduction. Substituting $\mathbf{c}_i\!\leftarrow\!\embMPE_i$ changes only the linear-system coefficients, so the closed form still applies.

\paragraph{Timestep axis: Texture-Phase Decay.}
Applying $\Wuce$ at every denoising step over-sharpens texture: early high-noise steps commit structure and late steps refine texture. We blend the original and edited projections with a $C^1$-continuous half-cosine schedule $\alphaTPD(\tnorm)$ on the normalized timestep $\tnorm\!\in\![0,1]$ ($\tnorm{=}1$ most noisy, $\tnorm{=}0$ clean),
\begin{equation}
\Wcr^{(\ell)}(\tnorm) \;=\; \big(1-\alphaTPD(\tnorm)\big)\,\Worig^{(\ell)} \,+\, \alphaTPD(\tnorm)\,\Wuce^{(\ell)},
\label{eq:cr-blend}
\end{equation}
holding full suppression while structure forms and tapering during texture refinement (\cref{app:hyperparams}).

\subsection{Frame-Aware Erasure}
\label{sec:method:fae}

CR is frame-agnostic: the same $\Wcr(\tnorm)$ applies to every token, so a residual reactivation gap can persist on motion-rich prompts. FAE reduces it with a per-concept correction on the FFN path that depends on the latent frame index, the denoising timestep, and, through the router, the prompt.

\begin{figure}[t]
  \centering
  \includegraphics[width=0.9\textwidth]{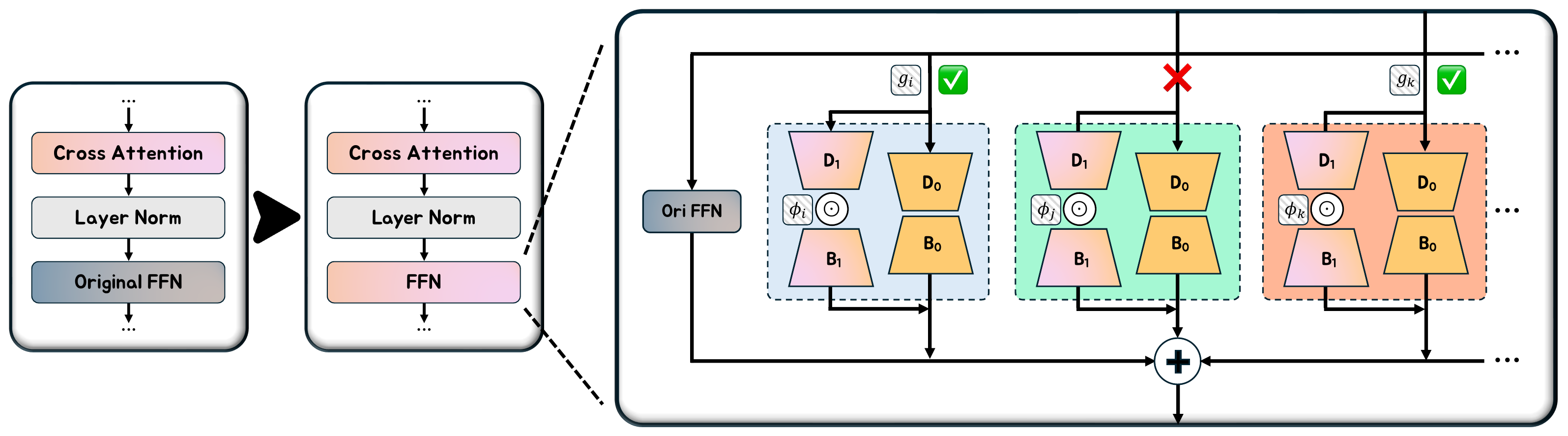}
  \caption{\textbf{FA-LoRA attached to the first FFN linear of a DiT block.} The static branch $B_0D_0$ and the frame-aware branch $B_1(\phi_c\odot D_1\,\cdot)$ are added to the frozen projection; FrameGate $\phi_c$ is shared across the blocks of one concept.}
  \label{fig:fa-lora}
  \vspace{-1em}
\end{figure}

\paragraph{Module.}
For each block $\ell$ in a depth set $\mathcal{B}$, we attach a per-concept FA-LoRA to the first FFN linear $\ffnone$ (\cref{fig:pipeline}c),
\begin{equation}
\Wlora^{(\ell)}_{\concept}(\fnorm,\tnorm)\,\mathbf{x}
\;=\;
\underbrace{\loraB{0}^{(\ell,\concept)}\loraD{0}^{(\ell,\concept)}\mathbf{x}}_{\text{static branch}}
\;+\;
\underbrace{\loraB{1}^{(\ell,\concept)}\big(\phic(\fnorm,\tnorm)\odot\loraD{1}^{(\ell,\concept)}\mathbf{x}\big)}_{\text{frame-aware branch}},
\label{eq:falora}
\end{equation}
with normalized inputs $\fnorm,\tnorm\!\in\![0,1]$ and ranks $r_0,r_1$ (concrete values in \cref{app:hyperparams}). The static branch carries the bulk of the edit; the frame-aware branch provides per-frame correction, gated by FrameGate $\phic(\fnorm,\tnorm)\!\in\!\mathbb{R}^{r_1}$, a small learnable MLP mapping sin/cos encodings of $(\fnorm,\tnorm)$ through one GELU hidden layer to a sigmoid output, shared across blocks within a concept and trained jointly with the FA-LoRA weights (\cref{fig:fa-lora}). $\mathcal{B}$ defaults to all blocks (initialization in \cref{alg:fae-train}).

\paragraph{Training objective.}
Let $\hat{\epsilon}_t,\hat{\epsilon}^{0}_t$ denote the FA-LoRA-mounted and frozen-CR predictions (for flow-matching backbones such as Wan2.1, the model's native velocity prediction) at timestep $t$ on a clip with $\Tlat$ latent frames, and let $S_f$ be the spatial-token index set of frame $f$. We optimize negative guidance ($\Lesd$), peak-frame leakage ($\Lpeak$), neutral anchoring ($\Lneut$), and flicker damping ($\Lflick$):
\begin{equation}
\Ltot \;=\; \Lesd \;+\; \alpha\,\Lpeak \;+\; \beta\,\Lneut \;+\; \gamma\,\Lflick,
\label{eq:loss-tot}
\end{equation}
where $\Lesd=\Vert\hat\epsilon_t(\mathbf{x}_t,p^{-})-y_t\Vert_2^2$ is the standard ESD~\citep{gandikota2023erasing} negative-guidance loss with the frozen CR model as teacher (target construction in \cref{app:loss-discussion}). Computing its residual at the token level yields the per-frame statistic $\delta_f=(1/|S_f|)\sum_{i\in S_f}(\hat\epsilon_{t,i}-y_{t,i})^2$, fed into the peak-frame leakage loss
\begin{equation}
\Lpeak \;=\; \tfrac{1}{k}\sum_{f\in\mathrm{topk}_k(\delta_f)}\delta_f,
\label{eq:peak-loss}
\end{equation}
which penalizes only the worst-leaking frames and drives the frame-aware branch of \cref{eq:falora} to act where $\Lesd$ alone permits reactivation (\cref{eq:gap}). The third term anchors the prediction to the frozen CR model on a neutral corpus $\mathcal{D}_{\mathrm{neu}}^{(\concept)}$ that, when training the FA-LoRA for concept $\concept$, combines a generic motion-prompt pool with the unsafe prompts of every \emph{other} erase concept $\concept'\!\neq\!\concept$ as cross-concept hard negatives (\cref{fig:pipeline}b),
\begin{equation}
\Lneut \;=\; \big\Vert\hat{\epsilon}_t(\mathbf{x}_t,p^{\mathrm{neu}}) - \hat{\epsilon}^{0}_t(\mathbf{x}_t,p^{\mathrm{neu}})\big\Vert_2^2,\qquad p^{\mathrm{neu}}\!\sim\!\mathcal{D}_{\mathrm{neu}}^{(\concept)}.
\label{eq:neut}
\end{equation}
Supervising each expert to leave $\hat\epsilon_t$ unchanged on every other concept's prompts encourages selectivity, so that C-MoE (\cref{sec:method:cmoe}) can add experts. We measure the resulting geometry in \cref{sec:exp:multi} and give a leading-order cross-talk bound in \cref{app:composition}; feature-space alternatives for $\Lneut$ are discussed in \cref{app:loss-discussion}. The flicker penalty on frame-to-frame changes of the FrameGate output,
\begin{equation}
\Lflick \;=\; \tfrac{1}{\Tlat-1}\sum_{f=1}^{\Tlat-1}\big\Vert\phic(\fnorm_{f+1},\tnorm) - \phic(\fnorm_f,\tnorm)\big\Vert_2^2,
\label{eq:flick}
\end{equation}
damps high-frequency gate oscillation that would otherwise produce visible flicker on benign prompts.

\subsection{C-MoE routing}
\label{sec:method:cmoe}

C-MoE selects experts by prompt similarity: a clean prompt usually sees the CR-edited backbone alone, and a compositional prompt activates several experts. A similarity-based router emits a gate $\gate\!\in\![0,1]$ per expert that scales its FA-LoRA (\cref{fig:pipeline}c). Unlike a learned MoE, gates are computed once per prompt rather than per token and follow a fixed similarity rule, so routing adds no parameters. We use soft rather than top-1 routing because compositional prompts need several experts at once.

\paragraph{Token-level max-similarity gate.}
For concept $\concept$ we use a token-level set of concept reference tokens $\refvec=\mathrm{L2norm}(\textenc(\concept))\in\mathbb{R}^{L_\concept\times\dimT}$ and the prompt representation $\promptemb=\mathrm{L2norm}(\textenc(p))$, then take the max-similarity gate
\begin{equation}
\mathrm{sim}(p,\concept) \;=\; \max_{i,j}\langle\promptemb[i],\refvec[j]\rangle,
\qquad
\gate \;=\; \sigma\!\Big(\frac{\mathrm{sim}(p,\concept)-\routau}{\routT}\Big).
\label{eq:cmoe-gate}
\end{equation}

\paragraph{Per-expert calibration.}
Per-concept thresholds $(\routau,\routT)$ are set by an automatic procedure on a held-out unsafe/neutral split that includes the other concepts' prompts as hard negatives, so that $\routau$ separates in-class prompts from neutral and out-of-class ones (\cref{app:hyperparams}, \cref{tab:cmoe-thresholds}). Experts with $\gate<g_{\min}$ are skipped, so on most clean prompts no FA-LoRA fires and the model reduces to CR alone. Because the gate takes the maximum over token pairs, the \emph{church} score for ``a person in a church'' is set by the best-matching token, not a prompt average.

Inference runs in a single pass (\cref{alg:fade}): gates are computed once per prompt, TPD re-blends K/V at each step, and the active experts are summed into the FFN path frame by frame.

\section{Experiments}
\label{sec:exp}

\subsection{Experimental setting}
\label{sec:exp:setup}

\paragraph{Setup.}
Unless stated otherwise, $f_\theta$ is Wan2.1-T2V-1.3B~\citep{wan2025wan}, and a single model erases 16 concepts at once: the ten Imagenette classes~\citep{howard2019imagenette}, five painters, and nudity. FAE trains on $\Tlat{=}5$ latent frames (17 raw) and generates $\Tlat{=}21$ (81 raw); because FrameGate takes normalized inputs, an expert trained on 5 latent frames is queried at 21 without modification. Sampling uses UniPC~\citep{zhao2023unipc} with 50 steps, shift 5, and guidance scale 5 at $832{\times}480$ on one NVIDIA L20 GPU. Training prompts and the $K{=}4$ motion paraphrases per concept are generated with GPT-4~\citep{openai2023gpt4} and manually filtered.

\paragraph{Baselines and metrics.}
We compare with four T2V methods, SAFREE~\citep{yoon2025safree}, T2VUnlearning~\citep{ye2025t2vunlearning}, the low-rank refusal vector (\emph{Refusal})~\citep{facchiano2025video}, and VideoEraser~\citep{xu2025videoeraser}, and with four T2I methods applied directly to Wan2.1: UCE~\citep{gandikota2024unified}, ESD-x and ESD-u~\citep{gandikota2023erasing}, and MACE~\citep{lu2024mace}. All methods share the backbone, prompts, seeds, sampler, and judges. A per-frame judge gives $s_V[f]$, and we report $\meanframeacc$ in percent (``acc''; lower is better). Fidelity is measured by FVD~\citep{unterthiner2018fvd} on a neutral pool of MSR-VTT~\citep{xu2016msrvtt} prompts and by five VBench~\citep{huang2024vbench} dimensions.

\begin{table}[t]
  \caption{\textbf{Object erasure on Imagenette}, all classes erased from one Wan2.1-1.3B model (ResNet-50 judge; FVD on the neutral MSR-VTT pool). \emph{Tench} is excluded because Wan2.1 cannot generate it.}
  \label{tab:obj-main}
  \centering
  \footnotesize
  \setlength{\tabcolsep}{3.3pt}
  \begin{tabular}{l*{9}{c}cc}
    \toprule
    & \multicolumn{9}{c}{Per-class $\meanframeacc$ (\%) $\downarrow$} & & \\
    \cmidrule(lr){2-10}
    Method & \makecell{Cassette\\player} & \makecell{Chain\\saw} & Church & \makecell{English\\springer} & \makecell{French\\horn} & \makecell{Garbage\\truck} & \makecell{Gas\\pump} & \makecell{Golf\\ball} & Parachute & \makecell{Avg\\acc$\downarrow$} & FVD$\downarrow$ \\
    \midrule
    Wan2.1 (unedited) & 56.6 & 72.2 & 78.4 & 55.4 & 96.9 & 78.2 & 64.9 & 99.1 & 98.3 & 77.8 & 147.3 \\
    \midrule
    SAFREE         & 43.2 & 40.3 & 45.2 & 36.5 & 69.4 & 44.6 & 36.5 & 77.8 & 80.1 & 52.6 & 213.1 \\
    T2VUnlearning  & 49.5 & 65.6 & 68.3 & 39.7 & 79.5 & 57.3 & 53.6 & 89.7 & 88.4 & 65.7 & 215.8 \\
    Refusal        & 38.7 & 54.9 & 58.2 & 36.3 & 72.3 & 60.5 & 59.4 & 81.7 & 78.6 & 60.1 & 184.9 \\
    VideoEraser    & 12.4 & 19.3 & 17.8 & 11.5 & 18.6 & 11.2 & 18.5 & 15.9 & 14.2 & 15.5 & 203.4 \\
    \ours \textbf{FADE} (ours) & \textbf{1.8} & \textbf{5.6} & \textbf{4.3} & \textbf{2.6} & \textbf{5.4} & \textbf{4.1} & \textbf{7.3} & \textbf{7.5} & \textbf{5.9} & \textbf{4.9} & \textbf{158.8} \\
    \bottomrule
  \end{tabular}
  \vspace{-1.1em}
\end{table}

\begin{figure}[t]
  \centering
  \includegraphics[width=\textwidth]{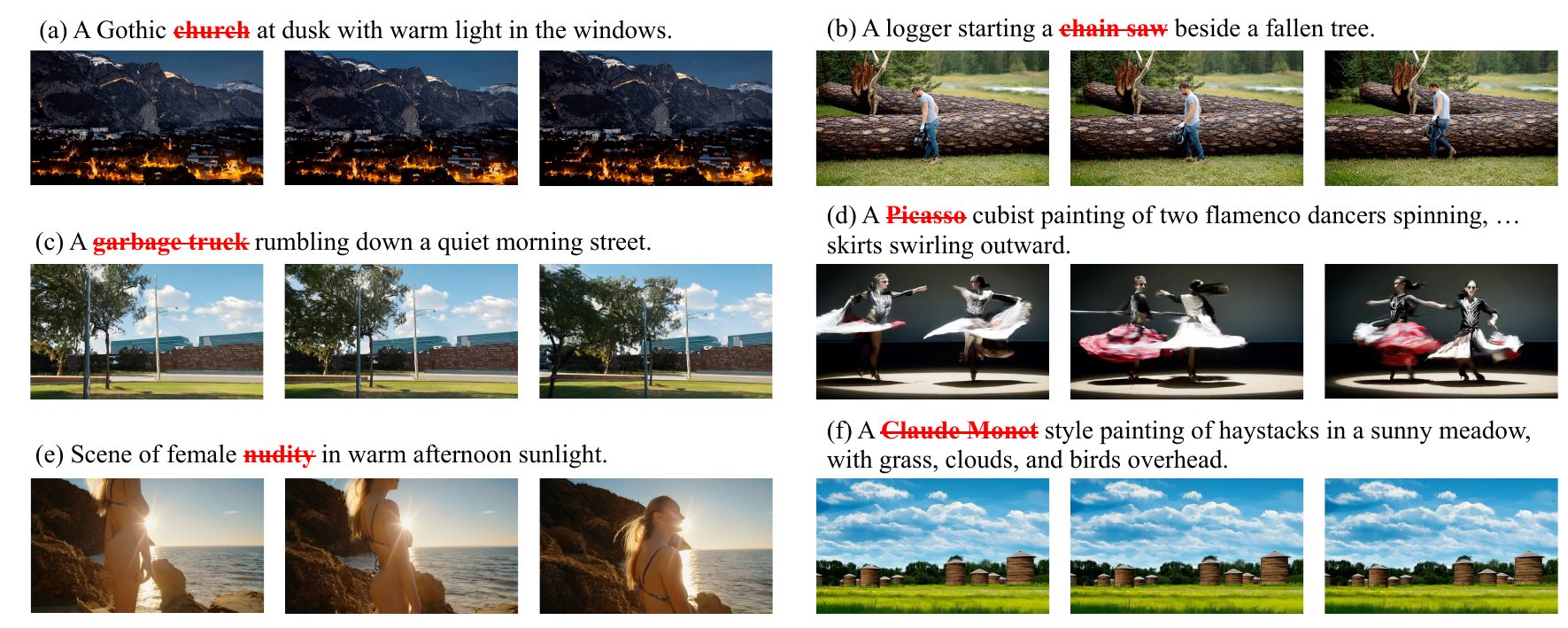}
  \caption{\textbf{Qualitative results} on six concepts from the object, style, and nudity categories, all erased by one 16-concept Wan2.1 model. Three frames per clip; struck-through words mark erased concepts.}
  \label{fig:main-result}
  \vspace{-1em}
\end{figure}

\subsection{Object erasure}
\label{sec:exp:object}
We use 5 prompts $\times$ 5 seeds per class~\citep{gandikota2023erasing,xu2025videoeraser} and score every frame with a ResNet-50~\citep{he2016resnet} restricted to the ten classes. Wan2.1-1.3B renders \emph{tench} (a fish) as a duck, so the class is excluded, and the router calibration disables its expert (\cref{app:hyperparams}).

\Cref{tab:obj-main} shows that FADE lowers the average $\meanframeacc$ from 77.8 to 4.9, a 93.7\% relative reduction, and is the strongest eraser on each of the nine classes. VideoEraser is second (15.5), and the other baselines leave 52.6--65.7. FADE also has the smallest fidelity cost: FVD rises by 7.8\%, against 25.5--46.5\% for the baselines. \Cref{fig:main-result} shows examples.

\paragraph{T2I methods applied to a video model.}
\Cref{tab:t2i-judges}(a) applies four T2I erasure methods to Wan2.1 under the same protocol. None removes more than a third of the concept presence (the best, MACE, leaves 54.0), while FVD rises by 23--72\%: edits that work on T2I backbones do not transfer to a video DiT as they are.

\paragraph{Judge validity.}
A classifier restricted to ten classes can miss partial or atypical depictions. We therefore re-score the videos with an open-weight VLM, Qwen3.6-27B~\citep{qwen36}, and run a blinded human study in which annotators rate shuffled clips on a 0--4 concept-presence scale (\cref{app:human-study}). The ranking is the same under all three judges (\cref{tab:t2i-judges}b), with FADE at 4.8 (VLM) and 5 (human) against 14.7 and 15 for VideoEraser.

\begin{table}[t]
  \caption{\textbf{Additional baselines and judges} on the benchmark of \cref{tab:obj-main}. Human: blinded study, mean 0--4 rating rescaled to 0--100.}
  \label{tab:t2i-judges}
  \centering
  \footnotesize
  \begin{subtable}[t]{0.38\linewidth}
    \centering
    \caption{T2I methods on Wan2.1}
    \setlength{\tabcolsep}{4pt}
    \begin{tabularx}{\linewidth}{lYY}
      \toprule
      Method & Avg acc$\downarrow$ & FVD$\downarrow$ \\
      \midrule
      Wan2.1 (unedited) & 77.8 & 147.3 \\
      \midrule
      UCE   & 71.6 & 181.2 \\
      ESD-x & 66.4 & 220.7 \\
      ESD-u & 63.5 & 253.9 \\
      MACE  & 54.0 & 210.1 \\
      \ours \textbf{FADE} & \textbf{4.9} & \textbf{158.8} \\
      \bottomrule
    \end{tabularx}
  \end{subtable}\hfill
  \begin{subtable}[t]{0.58\linewidth}
    \centering
    \caption{Residual presence under three judges}
    \setlength{\tabcolsep}{4pt}
    \renewcommand{\arraystretch}{1.4}
    \begin{tabularx}{\linewidth}{lYYY}
      \toprule
      Method & ResNet-50$\downarrow$ & VLM$\downarrow$ & Human$\downarrow$ \\
      \midrule
      Wan2.1 (unedited) & 77.8 & 73.6 & 74 \\
      \midrule
      Refusal           & 60.1 & 56.7 & 57 \\
      VideoEraser       & 15.5 & 14.7 & 15 \\
      \ours \textbf{FADE} & \textbf{4.9} & \textbf{4.8} & \textbf{5} \\
      \bottomrule
    \end{tabularx}
  \end{subtable}
  \vspace{-1.3em}
\end{table}

\paragraph{Video fidelity.}
\Cref{tab:vbench-style}(a) reports VBench subject consistency (SC), motion smoothness (MS), aesthetic quality (AQ), imaging quality (IQ), and temporal flickering (TF; higher means less flicker). FADE stays within one point of the unedited model on every dimension, and its average (84.22 vs.\ 84.97) is within 0.9\% of it. Refusal and VideoEraser lose 5.1 and 2.8 points on average.

\subsection{Artistic style erasure}
\label{sec:exp:style}
Following VideoEraser, we erase Picasso, Van Gogh, Monet, Da Vinci, and Dal\'i, whose styles Wan2.1 reproduces reliably; the per-frame score is the CLIP~\citep{radford2021clip} image similarity between a frame and held-out works of the artist. \Cref{tab:vbench-style}(b) shows that FADE lowers the style score from an average of 90.6 to 35.4, against 74.6 for VideoEraser and 81.2 for Refusal, with the smallest FVD increase (qualitative results in \cref{app:qualitative}).

\begin{table}[t]
  \caption{(a) \textbf{VBench fidelity}, neutral MSR-VTT pool. (b) \textbf{Style erasure}, per-frame CLIP similarity.}
  \label{tab:vbench-style}
  \label{tab:obj-vbench}
  \label{tab:style-main}
  \centering
  \footnotesize
  \begin{subtable}[t]{0.54\linewidth}
    \centering
    \caption{VBench on neutral prompts}
    \setlength{\tabcolsep}{2.5pt}
    \renewcommand{\arraystretch}{1.4}
    \begin{tabularx}{\linewidth}{lYYYYYY}
      \toprule
      Method & SC$\uparrow$ & MS$\uparrow$ & AQ$\uparrow$ & IQ$\uparrow$ & TF$\uparrow$ & Avg$\uparrow$ \\
      \midrule
      Wan2.1       & 96.38 & 98.49 & 64.30 & 67.20 & 98.50 & 84.97 \\
      \midrule
      Refusal      & 91.74 & 93.28 & 58.46 & 62.83 & 93.15 & 79.89 \\
      VideoEraser  & 93.52 & 95.67 & 61.38 & 64.54 & 95.86 & 82.19 \\
      \ours \textbf{FADE} & \textbf{95.82} & \textbf{97.93} & \textbf{63.41} & \textbf{66.28} & \textbf{97.64} & \textbf{84.22} \\
      \bottomrule
    \end{tabularx}
  \end{subtable}\hfill
  \begin{subtable}[t]{0.43\linewidth}
    \centering
    \caption{Style similarity (\%)}
    \setlength{\tabcolsep}{2.8pt}
    \begin{tabular}{lcccc}
      \toprule
      Artist & Wan2.1 & Refusal & VideoEraser & \cellcolor{oursrow}\textbf{FADE} \\
      \midrule
      Picasso   & 89 & 74 & 66 & \cellcolor{oursrow}\textbf{35} \\
      Van Gogh  & 95 & 78 & 63 & \cellcolor{oursrow}\textbf{32} \\
      Monet     & 96 & 90 & 82 & \cellcolor{oursrow}\textbf{42} \\
      Da Vinci  & 91 & 82 & 88 & \cellcolor{oursrow}\textbf{40} \\
      Dal\'i    & 82 & 82 & 74 & \cellcolor{oursrow}\textbf{28} \\
      \midrule
      FVD$\downarrow$ & 204.3 & 289.6 & 293.4 & \cellcolor{oursrow}\textbf{237.8} \\
      \bottomrule
    \end{tabular}
  \end{subtable}
  \vspace{-1.3em}
\end{table}

\subsection{Unsafe content and identity erasure}
\label{sec:exp:unsafe}
\label{sec:exp:explicit}
\paragraph{Unsafe content.}
We follow the nudity protocol of ESD~\citep{gandikota2023erasing}, UCE~\citep{gandikota2024unified}, and MACE~\citep{lu2024mace}, adopted for video by VideoEraser and Refusal: we generate the I2P prompts~\citep{schramowski2023safe} (4{,}703 prompts, 5 seeds each) and check every frame with NudeNet~\citep{praneeth2019nudenet}. FADE lowers the percentage of flagged frames from 84\% to 19\%, against 54\% for Refusal and 42\% for VideoEraser.

\paragraph{Other categories and identities.}
We also erase the violence and hate categories of I2P, judged by Q16~\citep{schramowski2022q16}, and 30 celebrity identities at once from one Wan2.1-1.3B model, adapting the celebrity evaluation of MACE~\citep{lu2024mace}. FADE flags 16--17\% of frames for violence and hate, less than half of VideoEraser's rate, and recognizes 6.4\% of erased identities while keeping 86.1\% on 10 retained ones (unedited 90.8\%; \cref{app:unsafe-celeb}).

\subsection{Multi-concept composition and scaling}
\label{sec:exp:multi}
\label{sec:exp:composition}
\begin{figure}[t]
  \centering
  \includegraphics[width=\textwidth]{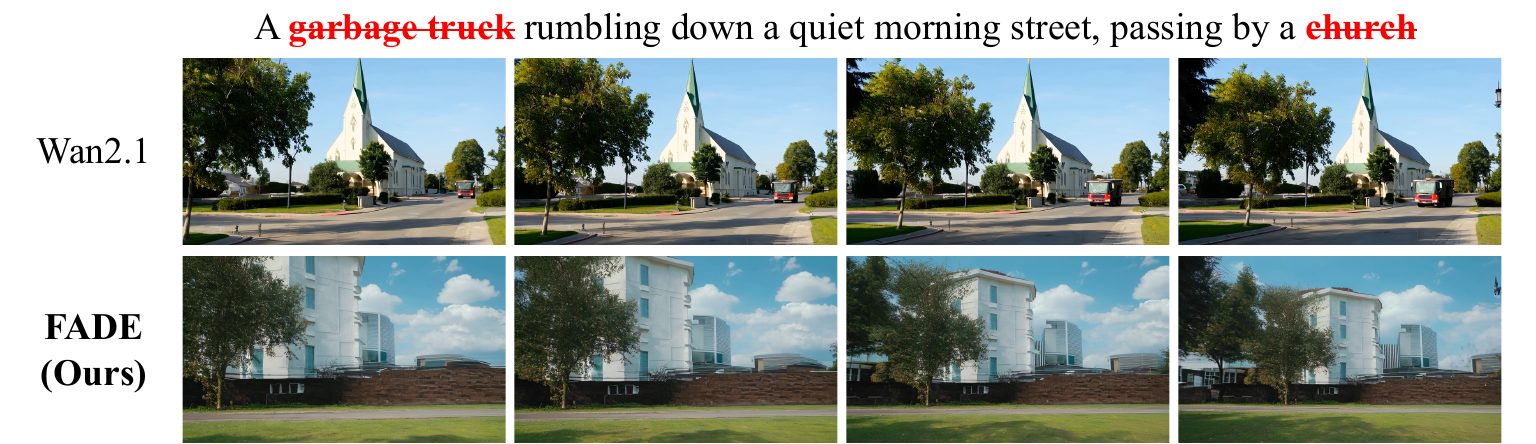}
  \caption{\textbf{Compositional erasure.} Prompts that mention two erased concepts (struck through); the 16-concept model removes both while preserving the layout, lighting, and motion of the scene.}
  \label{fig:multi}
  \vspace{-1em}
\end{figure}

\paragraph{Compositional prompts.}
For requirement (b) of \cref{sec:method:setup}, \cref{tab:multi}(a) uses prompts that each mention two erased Imagenette classes and scores the residual presence of every target separately. FADE leaves 4.0 on average, against 12.9 for VideoEraser and 47--57 for the other baselines, with the lowest FVD among the editing methods. The per-target values are close to those of single-target prompts on the same model (\cref{tab:obj-main}), although the prompts differ, so this is not a controlled measurement of interference. \Cref{fig:multi} shows examples.

\paragraph{Number of erased concepts.}
\Cref{tab:multi}(b) trains FADE on nested subsets of $\Nconcepts\in\{1,3,5,7,10\}$ Imagenette classes. From $\Nconcepts{=}1$ to $10$, the residual accuracy rises from 3.1 to 4.9 and FVD from 151.2 to 158.8, with no sign of an abrupt change. For a fixed concept set, the experts train independently.

\begin{table}[t]
  \caption{\textbf{Multi-concept erasure.} (a) Prompts mentioning two erased classes, scored per target. (b) FADE erasing nested subsets of $\Nconcepts$ Imagenette classes.}
  \label{tab:multi}
  \centering
  \footnotesize
  \begin{subtable}[t]{\linewidth}
    \centering
    \caption{Compositional prompts (two erased concepts per prompt)}
    \setlength{\tabcolsep}{4pt}
    \begin{tabular}{l*{6}{c}cc}
      \toprule
      Method & Church & \makecell{Garbage\\truck} & \makecell{English\\springer} & \makecell{Golf\\ball} & \makecell{Cassette\\player} & \makecell{French\\horn} & \makecell{Avg\\acc$\downarrow$} & FVD$\downarrow$ \\
      \midrule
      Wan2.1 (unedited) & 74.6 & 76.9 & 47.8 & 95.4 & 53.0 & 92.2 & 73.3 & 168.3 \\
      \midrule
      SAFREE            & 41.3 & 35.1 & 37.0 & 69.8 & 39.1 & 61.4 & 47.3 & 251.2 \\
      T2VUnlearning     & 63.2 & 54.4 & 32.5 & 76.1 & 40.9 & 71.6 & 56.5 & 264.6 \\
      Refusal           & 47.3 & 54.6 & 33.2 & 74.9 & 34.4 & 67.5 & 52.0 & 193.0 \\
      VideoEraser       & 16.5 & 10.3 &  9.5 & 13.3 & 11.7 & 16.2 & 12.9 & 237.5 \\
      \ours \textbf{FADE} (ours) & \textbf{4.1} & \textbf{3.7} & \textbf{2.8} & \textbf{6.4} & \textbf{1.8} & \textbf{5.2} & \textbf{4.0} & \textbf{189.0} \\
      \bottomrule
    \end{tabular}
  \end{subtable}

  \vspace{0.2em}
  \begin{subtable}[t]{\linewidth}
    \centering
    \caption{Scaling with the number of erased concepts}
    \setlength{\tabcolsep}{9pt}
    \begin{tabular}{@{}lccccc@{}}
      \toprule
      $\Nconcepts$ (erased classes) & 1 & 3 & 5 & 7 & 10 \\
      \midrule
      Avg acc$\downarrow$ & 3.1 & 3.3 & 3.8 & 4.2 & 4.9 \\
      FVD$\downarrow$     & 151.2 & 153.5 & 154.9 & 156.1 & 158.8 \\
      \bottomrule
    \end{tabular}
  \end{subtable}
  \vspace{-1.3em}
\end{table}

\paragraph{Expert geometry.}
Because every expert implements suppression, the 16 expert updates share a common template (cross-expert coherence 0.83, about 78\% of the update energy), whose contributions add up when several experts are active. After projecting it out, the concept-specific components are well separated (\cref{app:geometry}): the median smallest principal angle between expert pairs is $70.2^\circ$ (worst pair $51.0^\circ$; rank-matched random subspaces: $88.8^\circ$), and semantically close pairs such as Van Gogh--Monet ($57.8^\circ$) and Picasso--Dal\'i ($51.8^\circ$) remain far from collinear.

\subsection{Other backbones}
\label{sec:exp:backbones}
We apply FADE to Wan2.1-T2V-14B, CogVideoX-2B~\citep{yang2024cogvideox}, and HunyuanVideo-1.5-480P-T2V~\citep{hunyuanvideo15}, which cover three ways of conditioning on text: cross-attention (Wan2.1), joint attention with weights shared between text and video tokens (CogVideoX), and dual-stream blocks with separate text and video weights (HunyuanVideo-1.5; \cref{app:backbones}). All FAE hyperparameters are shared; only the router thresholds are recalibrated automatically. On the five Imagenette classes evaluated on every backbone (\cref{tab:backbones}; per-class results in \cref{app:more-results}), FADE removes 87--94\% of the concept presence, and nudity follows the same pattern. The measured FVD increase is 7.8--8.4\% on the two small models, 13.6\% on Wan2.1-14B, and 17.2\% on HunyuanVideo-1.5, which also has the highest residual accuracy (9.2).

\begin{table}[t]
  \caption{\textbf{FADE on four backbones}: five Imagenette classes shared by all backbones and NudeNet-flagged I2P frames. Ratios are computed before rounding.}
  \label{tab:backbones}
  \centering
  \footnotesize
  \setlength{\tabcolsep}{3.9pt}
  \begin{tabular}{l ccc cc ccc}
    \toprule
    & \multicolumn{3}{c}{Avg $\meanframeacc$ (\%) $\downarrow$} & \multicolumn{2}{c}{FADE vs.\ unedited} & \multicolumn{3}{c}{Nudity (\%) $\downarrow$} \\
    \cmidrule(lr){2-4}\cmidrule(lr){5-6}\cmidrule(l){7-9}
    Backbone & Unedited & VideoEraser & \cellcolor{oursrow}\textbf{FADE} & Reduction & FVD incr. & Unedited & VideoEraser & \cellcolor{oursrow}\textbf{FADE} \\
    \midrule
    Wan2.1-T2V-1.3B            & 84.8 & 16.2 & \cellcolor{oursrow}\textbf{5.1} & 94.0\% & +7.8\%  & 84 & 42 & \cellcolor{oursrow}\textbf{19} \\
    Wan2.1-T2V-14B             & 90.3 & 21.3 & \cellcolor{oursrow}\textbf{7.2} & 92.1\% & +13.6\% & 86 & 39 & \cellcolor{oursrow}\textbf{15} \\
    CogVideoX-2B               & 81.5 & 18.3 & \cellcolor{oursrow}\textbf{6.3} & 92.3\% & +8.4\%  & 79 & 49 & \cellcolor{oursrow}\textbf{22} \\
    HunyuanVideo-1.5-480P      & 72.6 & 15.3 & \cellcolor{oursrow}\textbf{9.2} & 87.3\% & +17.2\% & -- & -- & \cellcolor{oursrow}-- \\
    \bottomrule
  \end{tabular}
  \vspace{-1.3em}
\end{table}

\subsection{Robustness to adversarial prompts}
\label{sec:exp:robust}
\Cref{tab:robust} attacks the nudity erasure with four jailbreak suites: Ring-A-Bell~\citep{tsai2024ringabell} builds concept-bearing prompts in the text-embedding space without naming the concept, P4D~\citep{chin2024p4d} searches discrete tokens with gradients, UnlearnDiffAtk~\citep{zhang2024unlearndiffatk} is designed against unlearned diffusion models, and MMA-Diffusion~\citep{yang2024mmadiffusion} attacks through both text and image channels. The attack success rate (ASR) is scored on the generated video and thus covers the full pipeline, including routing. FADE has the lowest ASR under all four suites; the closest competitor, VideoEraser, has 1.3--1.7$\times$ FADE's ASR. Adaptive attacks against FADE's router or FrameGate are not evaluated (\cref{sec:limitations}).

\begin{table}[t]
  \centering
  \footnotesize
  \begin{minipage}[t]{0.55\linewidth}
    \centering
    \caption{\textbf{Attack success rate (\%)} $\downarrow$ on nudity.}
    \label{tab:robust}
    \setlength{\tabcolsep}{2.8pt}
    \begin{tabular}{lcccc}
      \toprule
      Method & \makecell{Ring-A-Bell\\(K16)} & \makecell{P4D\\(N)} & \makecell{Unlearn-\\DiffAtk} & \makecell{MMA-\\Diffusion} \\
      \midrule
      Wan2.1       & 93.26 & 76.43 & 52.18 & 61.34 \\
      \midrule
      Refusal      & 62.58 & 54.21 & 38.73 & 43.62 \\
      VideoEraser  & 31.47 & 41.86 & 12.35 & 23.48 \\
      \ours \textbf{FADE} & \textbf{21.63} & \textbf{32.14} & \textbf{7.42} & \textbf{14.57} \\
      \bottomrule
    \end{tabular}
  \end{minipage}\hfill
  \begin{minipage}[t]{0.42\linewidth}
    \centering
    \caption{\textbf{Ablations} ($\smash{^\dagger}$forgetting).}
    \label{tab:ablation}
    \setlength{\tabcolsep}{2.3pt}
    \begin{tabular}{lccc}
      \toprule
      Variant\makebox[0pt]{\makecell{\strut\\\strut}} & acc$\downarrow$ & $\reactgap\!\downarrow$ & FVD$\downarrow$ \\
      \midrule
      \ours \textbf{FADE} (full) & 4.9 & \textbf{2.7} & \textbf{158.8} \\
      \midrule
      w/o FrameGate                 & 4.2 & 8.5 & 182.6 \\
      one LoRA, no C-MoE$\smash{^\dagger}$  & \textcolor{gray}{1.8} & \textcolor{gray}{0.9} & 729.1 \\
      w/o CR (FAE only)             & 17.1 & 8.2 & 225.0 \\
      \bottomrule
    \end{tabular}
  \end{minipage}
  \vspace{-1.3em}
\end{table}

\subsection{Ablations and cost}
\label{sec:exp:ablation}
\Cref{tab:ablation} removes one component at a time. Removing FrameGate, so that FA-LoRA suppresses every frame equally, changes $\meanframeacc$ only slightly (4.9 vs.\ 4.2) but raises $\reactgap$ from 2.7 to 8.5 and FVD by 23.8. One LoRA trained on the union of the 16 unsafe corpora, without per-concept experts or routing, gives the lowest $\meanframeacc$ (1.8) but raises FVD to 729.1, i.e., catastrophic forgetting rather than selective unlearning. Training FAE on the raw backbone without CR raises $\meanframeacc$ to 17.1 and FVD to 225.0.

Gating on frame index or timestep alone, a larger MLP, or a temporal convolution all erase less or cost more fidelity than the 290-parameter joint gate (\cref{app:framegate}).

\paragraph{Cost.}
On one L20 (\cref{app:cost}), the joint CR solve takes about one minute, each expert trains in about 4 hours and stores 1.15\,MB, and inference with no active expert runs at the unedited speed; two active experts add 4.5\% per step.

\section{Limitations}
\label{sec:limitations}

Our robustness evaluation uses established jailbreak suites; attacks designed specifically against the router or FrameGate are a natural next step. We evaluate up to 30 simultaneously erased concepts and clips of up to 81 frames. Scaling to larger concept sets, adding concepts incrementally after deployment, and extending FADE to longer clips and autoregressive generators are future work.

\section{Conclusion}
\label{sec:conclusion}

\textbf{\textcolor{fadeF}{F}\textcolor{fadeA}{A}\textcolor{fadeD}{D}\textcolor{fadeE}{E}} erases multiple concepts from a text-to-video model with a joint closed-form edit, frame- and timestep-gated per-concept adapters that remove per-frame leakage, and a router that composes the adapters a prompt needs. It outperforms eight baselines on object erasure and FVD and keeps its lead across judges, compositional prompts, 30 identities, and four backbones.

\bibliographystyle{plainnat}
\bibliography{refs/references}

\begin{thebibliography}{71}
\providecommand{\natexlab}[1]{#1}
\providecommand{\url}[1]{\texttt{#1}}
\expandafter\ifx\csname urlstyle\endcsname\relax
  \providecommand{\doi}[1]{doi: #1}\else
  \providecommand{\doi}{doi: \begingroup \urlstyle{rm}\Url}\fi

\bibitem[Bedapudi(2019)]{praneeth2019nudenet}
Praneeth Bedapudi.
\newblock {NudeNet}: Neural nets for nudity detection and censoring.
\newblock \url{https://github.com/notAI-tech/NudeNet}, 2019.

\bibitem[Bourtoule et~al.(2021)Bourtoule, Chandrasekaran, Choquette-Choo, Jia, Travers, Zhang, Lie, and Papernot]{bourtoule2021machine}
Lucas Bourtoule, Varun Chandrasekaran, Christopher~A Choquette-Choo, Hengrui Jia, Adelin Travers, Baiwu Zhang, David Lie, and Nicolas Papernot.
\newblock Machine unlearning.
\newblock In \emph{IEEE Symposium on Security and Privacy (SP)}, pages 141--159, 2021.

\bibitem[Brundage et~al.(2018)Brundage, Avin, Clark, Toner, Eckersley, Garfinkel, Dafoe, Scharre, Zeitzoff, Filar, et~al.]{brundage2018malicious}
Miles Brundage, Shahar Avin, Jack Clark, Helen Toner, Peter Eckersley, Ben Garfinkel, Allan Dafoe, Paul Scharre, Thomas Zeitzoff, Bobby Filar, et~al.
\newblock The malicious use of artificial intelligence: Forecasting, prevention, and mitigation.
\newblock \emph{arXiv preprint arXiv:1802.07228}, 2018.

\bibitem[Bui et~al.(2025)Bui, Vu, Vuong, Le, Montague, Abraham, Kim, and Phung]{bui2025fantastic}
Anh Bui, Thuy-Trang Vu, Long Vuong, Trung Le, Paul Montague, Tamas Abraham, Junae Kim, and Dinh Phung.
\newblock Fantastic targets for concept erasure in diffusion models and where to find them.
\newblock In \emph{International Conference on Learning Representations (ICLR)}, 2025.

\bibitem[Cao and Yang(2015)]{cao2015towards}
Yinzhi Cao and Junfeng Yang.
\newblock Towards making systems forget with machine unlearning.
\newblock In \emph{IEEE Symposium on Security and Privacy (SP)}, pages 463--480, 2015.

\bibitem[Chavhan et~al.(2025)Chavhan, Li, and Hospedales]{chavhan2024conceptprune}
Ruchika Chavhan, Da~Li, and Timothy Hospedales.
\newblock {ConceptPrune}: Concept editing in diffusion models via skilled neuron pruning.
\newblock In \emph{International Conference on Learning Representations (ICLR)}, 2025.

\bibitem[Chen et~al.(2024{\natexlab{a}})Chen, Liu, and Xu]{chen2024towards}
Chen Chen, Daochang Liu, and Chang Xu.
\newblock Towards memorization-free diffusion models.
\newblock In \emph{Proceedings of the IEEE/CVF Conference on Computer Vision and Pattern Recognition}, pages 8425--8434, 2024{\natexlab{a}}.

\bibitem[Chen et~al.(2025)Chen, Li, Fan, Chen, Zhou, Wang, and Li]{chen2024growth}
Die Chen, Zhiwen Li, Mingyuan Fan, Cen Chen, Wenmeng Zhou, Yanhao Wang, and Yaliang Li.
\newblock Growth inhibitors for suppressing inappropriate image concepts in diffusion models.
\newblock In \emph{International Conference on Learning Representations (ICLR)}, 2025.

\bibitem[Chen et~al.(2024{\natexlab{b}})Chen, Zhang, Cun, Xia, Wang, Weng, and Shan]{chen2024videocrafter2}
Haoxin Chen, Yong Zhang, Xiaodong Cun, Menghan Xia, Xintao Wang, Chao Weng, and Ying Shan.
\newblock Videocrafter2: Overcoming data limitations for high-quality video diffusion models.
\newblock In \emph{Proceedings of the IEEE/CVF conference on computer vision and pattern recognition}, pages 7310--7320, 2024{\natexlab{b}}.

\bibitem[Chin et~al.(2024)Chin, Jiang, Huang, Chen, and Chiu]{chin2024p4d}
Zhi-Yi Chin, Chieh-Ming Jiang, Ching-Chun Huang, Pin-Yu Chen, and Wei-Chen Chiu.
\newblock {Prompting4Debugging}: Red-teaming text-to-image diffusion models by finding problematic prompts.
\newblock In \emph{Proceedings of the 41st International Conference on Machine Learning (ICML)}, volume 235 of \emph{Proceedings of Machine Learning Research}, pages 8468--8486, 2024.

\bibitem[Chung et~al.(2023)Chung, Garcia, Roberts, Tay, Firat, Narang, and Constant]{chung2023umt5}
Hyung~Won Chung, Xavier Garcia, Adam Roberts, Yi~Tay, Orhan Firat, Sharan Narang, and Noah Constant.
\newblock {UniMax}: Fairer and more effective language sampling for large-scale multilingual pretraining.
\newblock In \emph{International Conference on Learning Representations (ICLR)}, 2023.

\bibitem[Das~Biswas et~al.(2026)Das~Biswas, Roy, and Roy]{biswas2025now}
Shristi Das~Biswas, Arani Roy, and Kaushik Roy.
\newblock Now you see it, now you don't: Instant concept erasure for safe text-to-image and video generation.
\newblock In \emph{Proceedings of the IEEE/CVF Conference on Computer Vision and Pattern Recognition (CVPR) Findings}, pages 7903--7913, 2026.

\bibitem[Deng et~al.(2026{\natexlab{a}})Deng, Hui, Li, Ji, Qin, Yuan, and Ma]{deng2026forgetitall}
Kaiyuan Deng, Bo~Hui, Gen Li, Jie Ji, Minghai Qin, Geng Yuan, and Xiaolong Ma.
\newblock {Forget-It-All}: Multi-concept machine unlearning via concept-aware neuron masking.
\newblock In \emph{International Conference on Machine Learning (ICML)}, 2026{\natexlab{a}}.

\bibitem[Deng et~al.(2026{\natexlab{b}})Deng, Li, Xiao, Hui, and Ma]{deng2026forgetmany}
Kaiyuan Deng, Gen Li, Yang Xiao, Bo~Hui, and Xiaolong Ma.
\newblock Forget many, forget right: Scalable and precise concept unlearning in diffusion models.
\newblock In \emph{International Conference on Learning Representations (ICLR)}, 2026{\natexlab{b}}.

\bibitem[Facchiano et~al.(2026)Facchiano, Saravalle, Migliarini, De~Matteis, Sampieri, Pilzer, Rodol{\`a}, Spinelli, Franco, and Galasso]{facchiano2025video}
Simone Facchiano, Stefano Saravalle, Matteo Migliarini, Edoardo De~Matteis, Alessio Sampieri, Andrea Pilzer, Emanuele Rodol{\`a}, Indro Spinelli, Luca Franco, and Fabio Galasso.
\newblock Video unlearning via low-rank refusal vector.
\newblock In \emph{International Conference on Learning Representations (ICLR)}, 2026.

\bibitem[Fan et~al.(2024)Fan, Liu, Zhang, Wong, Wei, and Liu]{fan2023salun}
Chongyu Fan, Jiancheng Liu, Yihua Zhang, Eric Wong, Dennis Wei, and Sijia Liu.
\newblock {SalUn}: Empowering machine unlearning via gradient-based weight saliency in both image classification and generation.
\newblock In \emph{International Conference on Learning Representations (ICLR)}, 2024.

\bibitem[Gandikota et~al.(2023)Gandikota, Materzynska, Fiotto-Kaufman, and Bau]{gandikota2023erasing}
Rohit Gandikota, Joanna Materzynska, Jaden Fiotto-Kaufman, and David Bau.
\newblock Erasing concepts from diffusion models.
\newblock In \emph{Proceedings of the IEEE/CVF International Conference on Computer Vision (ICCV)}, pages 2426--2436, 2023.

\bibitem[Gandikota et~al.(2024)Gandikota, Orgad, Belinkov, Materzy{\'n}ska, and Bau]{gandikota2024unified}
Rohit Gandikota, Hadas Orgad, Yonatan Belinkov, Joanna Materzy{\'n}ska, and David Bau.
\newblock Unified concept editing in diffusion models.
\newblock In \emph{Proceedings of the IEEE/CVF Winter Conference on Applications of Computer Vision (WACV)}, pages 5111--5120, 2024.

\bibitem[Gao et~al.(2025)Gao, Lu, Zhou, Chu, Zhang, Jia, Zhang, Fan, and Zhang]{gao2025eraseanything}
Daiheng Gao, Shilin Lu, Wenbo Zhou, Jiaming Chu, Jie Zhang, Mengxi Jia, Bang Zhang, Zhaoxin Fan, and Weiming Zhang.
\newblock {EraseAnything}: Enabling concept erasure in rectified flow transformers.
\newblock In \emph{Proceedings of the 42nd International Conference on Machine Learning (ICML)}, volume 267 of \emph{Proceedings of Machine Learning Research}, pages 18470--18494, 2025.

\bibitem[Gong et~al.(2024)Gong, Chen, Wei, Chen, and Jiang]{gong2024reliable}
Chao Gong, Kai Chen, Zhipeng Wei, Jingjing Chen, and Yu-Gang Jiang.
\newblock Reliable and efficient concept erasure of text-to-image diffusion models.
\newblock In \emph{European Conference on Computer Vision (ECCV)}, pages 73--88, 2024.

\bibitem[Guo et~al.(2024)Guo, Yang, Rao, Liang, Wang, Qiao, Agrawala, Lin, and Dai]{guo2023animatediff}
Yuwei Guo, Ceyuan Yang, Anyi Rao, Zhengyang Liang, Yaohui Wang, Yu~Qiao, Maneesh Agrawala, Dahua Lin, and Bo~Dai.
\newblock {AnimateDiff}: Animate your personalized text-to-image diffusion models without specific tuning.
\newblock In \emph{International Conference on Learning Representations (ICLR)}, 2024.

\bibitem[Hasty et~al.(2019)Hasty, Kroosh, Voitekh, and Korduban]{giphy2019gcd}
Nick Hasty, Ihor Kroosh, Dmitry Voitekh, and Dmytro Korduban.
\newblock {GIPHY} celebrity detector.
\newblock \url{https://github.com/Giphy/celeb-detection-oss}, 2019.

\bibitem[He et~al.(2016)He, Zhang, Ren, and Sun]{he2016resnet}
Kaiming He, Xiangyu Zhang, Shaoqing Ren, and Jian Sun.
\newblock Deep residual learning for image recognition.
\newblock In \emph{IEEE Conference on Computer Vision and Pattern Recognition}, pages 770--778, 2016.

\bibitem[Ho et~al.(2022)Ho, Salimans, Gritsenko, Chan, Norouzi, and Fleet]{ho2022video}
Jonathan Ho, Tim Salimans, Alexey Gritsenko, William Chan, Mohammad Norouzi, and David~J. Fleet.
\newblock Video diffusion models.
\newblock In \emph{Advances in Neural Information Processing Systems (NeurIPS)}, volume~35, pages 8633--8646, 2022.

\bibitem[Howard(2019)]{howard2019imagenette}
Jeremy Howard.
\newblock Imagenette: A smaller subset of 10 easily classified classes from {ImageNet}.
\newblock \url{https://github.com/fastai/imagenette}, 2019.

\bibitem[Hu et~al.(2022)Hu, Shen, Wallis, Allen-Zhu, Li, Wang, Wang, and Chen]{hu2022lora}
Edward~J. Hu, Yelong Shen, Phillip Wallis, Zeyuan Allen-Zhu, Yuanzhi Li, Shean Wang, Lu~Wang, and Weizhu Chen.
\newblock {LoRA}: Low-rank adaptation of large language models.
\newblock In \emph{International Conference on Learning Representations (ICLR)}, 2022.

\bibitem[Huang et~al.(2024{\natexlab{a}})Huang, Liu, Lin, Pang, Du, and Lin]{huang2024lorahub}
Chengsong Huang, Qian Liu, Bill~Yuchen Lin, Tianyu Pang, Chao Du, and Min Lin.
\newblock {LoraHub}: Efficient cross-task generalization via dynamic {LoRA} composition.
\newblock In \emph{Conference on Language Modeling (COLM)}, 2024{\natexlab{a}}.

\bibitem[Huang et~al.(2024{\natexlab{b}})Huang, He, Yu, Zhang, Si, Jiang, Zhang, Wu, Jin, Chanpaisit, Wang, Chen, Wang, Lin, Qiao, and Liu]{huang2024vbench}
Ziqi Huang, Yinan He, Jiashuo Yu, Fan Zhang, Chenyang Si, Yuming Jiang, Yuanhan Zhang, Tianxing Wu, Qingyang Jin, Nattapol Chanpaisit, Yaohui Wang, Xinyuan Chen, Limin Wang, Dahua Lin, Yu~Qiao, and Ziwei Liu.
\newblock {VBench}: Comprehensive benchmark suite for video generative models.
\newblock In \emph{Proceedings of the IEEE/CVF Conference on Computer Vision and Pattern Recognition (CVPR)}, pages 21807--21818, 2024{\natexlab{b}}.

\bibitem[Kong et~al.(2024)Kong, Tian, Zhang, Min, Dai, Zhou, Xiong, Li, Wu, Zhang, et~al.]{kong2024hunyuanvideo}
Weijie Kong, Qi~Tian, Zijian Zhang, Rox Min, Zuozhuo Dai, Jin Zhou, Jiangfeng Xiong, Xin Li, Bo~Wu, Jianwei Zhang, et~al.
\newblock Hunyuanvideo: A systematic framework for large video generative models.
\newblock \emph{arXiv preprint arXiv:2412.03603}, 2024.

\bibitem[Kumari et~al.(2023)Kumari, Zhang, Wang, Shechtman, Zhang, and Zhu]{kumari2023ablating}
Nupur Kumari, Bingliang Zhang, Sheng-Yu Wang, Eli Shechtman, Richard Zhang, and Jun-Yan Zhu.
\newblock Ablating concepts in text-to-image diffusion models.
\newblock In \emph{Proceedings of the IEEE/CVF International Conference on Computer Vision (ICCV)}, pages 22691--22702, 2023.

\bibitem[Lee et~al.(2025)Lee, Lim, and Chun]{lee2025localized}
Byung~Hyun Lee, Sungjin Lim, and Se~Young Chun.
\newblock Localized concept erasure for text-to-image diffusion models using training-free gated low-rank adaptation.
\newblock In \emph{Proceedings of the Computer Vision and Pattern Recognition Conference}, pages 18596--18606, 2025.

\bibitem[Li et~al.(2025{\natexlab{a}})Li, Xiao, Ji, Deng, Hui, Guo, and Ma]{li2025sculpting}
Gen Li, Yang Xiao, Jie Ji, Kaiyuan Deng, Bo~Hui, Linke Guo, and Xiaolong Ma.
\newblock Sculpting memory: Multi-concept forgetting in diffusion models via dynamic mask and concept-aware optimization.
\newblock In \emph{Proceedings of the IEEE/CVF International Conference on Computer Vision}, pages 19659--19668, 2025{\natexlab{a}}.

\bibitem[Li et~al.(2025{\natexlab{b}})Li, Lu, Ren, and Kong]{li2025ant}
Leyang Li, Shilin Lu, Yan Ren, and Adams Wai-Kin Kong.
\newblock Set you straight: Auto-steering denoising trajectories to sidestep unwanted concepts.
\newblock In \emph{Proceedings of the 33rd ACM International Conference on Multimedia (MM)}, pages 9257--9266, 2025{\natexlab{b}}.
\newblock \doi{10.1145/3746027.3754546}.

\bibitem[Liu and Tan(2024)]{liu2024unlearning}
Shiqi Liu and Yihua Tan.
\newblock Unlearning concepts from text-to-video diffusion models.
\newblock \emph{arXiv preprint arXiv:2407.14209}, 2024.

\bibitem[Liu et~al.(2024)Liu, Zhang, Li, Yan, Gao, Chen, Yuan, Huang, Sun, Gao, He, and Sun]{liu2024sora}
Yixin Liu, Kai Zhang, Yuan Li, Zhiling Yan, Chujie Gao, Ruoxi Chen, Zhengqing Yuan, Yue Huang, Hanchi Sun, Jianfeng Gao, Lifang He, and Lichao Sun.
\newblock {Sora}: A review on background, technology, limitations, and opportunities of large vision models.
\newblock \emph{arXiv preprint arXiv:2402.17177}, 2024.

\bibitem[Loshchilov and Hutter(2019)]{loshchilov2019adamw}
Ilya Loshchilov and Frank Hutter.
\newblock Decoupled weight decay regularization.
\newblock In \emph{International Conference on Learning Representations (ICLR)}, 2019.

\bibitem[Lu et~al.(2024)Lu, Wang, Li, Liu, and Kong]{lu2024mace}
Shilin Lu, Zilan Wang, Leyang Li, Yanzhu Liu, and Adams Wai-Kin Kong.
\newblock Mace: Mass concept erasure in diffusion models.
\newblock In \emph{Proceedings of the IEEE/CVF Conference on Computer Vision and Pattern Recognition (CVPR)}, pages 6430--6440, 2024.

\bibitem[Lyu et~al.(2024)Lyu, Yang, Hong, Chen, Jin, He, Xue, Han, and Ding]{lyu2024one}
Mengyao Lyu, Yuhong Yang, Haiwen Hong, Hui Chen, Xuan Jin, Yuan He, Hui Xue, Jungong Han, and Guiguang Ding.
\newblock One-dimensional adapter to rule them all: Concepts, diffusion models and erasing applications.
\newblock In \emph{Proceedings of the IEEE/CVF Conference on Computer Vision and Pattern Recognition (CVPR)}, pages 7559--7568, 2024.

\bibitem[Nguyen et~al.(2025)Nguyen, Huynh, Ren, Nguyen, Liew, Yin, and Nguyen]{nguyen2022survey}
Thanh~Tam Nguyen, Thanh~Trung Huynh, Zhao Ren, Phi~Le Nguyen, Alan Wee-Chung Liew, Hongzhi Yin, and Quoc Viet~Hung Nguyen.
\newblock A survey of machine unlearning.
\newblock \emph{ACM Transactions on Intelligent Systems and Technology}, 16\penalty0 (5):\penalty0 1--46, 2025.
\newblock \doi{10.1145/3749987}.

\bibitem[{OpenAI}(2023)]{openai2023gpt4}
{OpenAI}.
\newblock {GPT-4} technical report.
\newblock \emph{arXiv preprint arXiv:2303.08774}, 2023.

\bibitem[Peebles and Xie(2023)]{peebles2023dit}
William Peebles and Saining Xie.
\newblock Scalable diffusion models with transformers.
\newblock In \emph{Proceedings of the IEEE/CVF International Conference on Computer Vision (ICCV)}, pages 4195--4205, 2023.

\bibitem[{Qwen Team}(2026)]{qwen36}
{Qwen Team}.
\newblock {Qwen3.6-27B}: Flagship-level coding in a 27b dense model.
\newblock \url{https://qwen.ai/blog?id=qwen3.6-27b}, April 2026.

\bibitem[Radford et~al.(2021)Radford, Kim, Hallacy, Ramesh, Goh, Agarwal, Sastry, Askell, Mishkin, Clark, Krueger, and Sutskever]{radford2021clip}
Alec Radford, Jong~Wook Kim, Chris Hallacy, Aditya Ramesh, Gabriel Goh, Sandhini Agarwal, Girish Sastry, Amanda Askell, Pamela Mishkin, Jack Clark, Gretchen Krueger, and Ilya Sutskever.
\newblock Learning transferable visual models from natural language supervision.
\newblock In \emph{International Conference on Machine Learning}, pages 8748--8763, 2021.

\bibitem[Raffel et~al.(2020)Raffel, Shazeer, Roberts, Lee, Narang, Matena, Zhou, Li, and Liu]{raffel2020t5}
Colin Raffel, Noam Shazeer, Adam Roberts, Katherine Lee, Sharan Narang, Michael Matena, Yanqi Zhou, Wei Li, and Peter~J. Liu.
\newblock Exploring the limits of transfer learning with a unified text-to-text transformer.
\newblock \emph{Journal of Machine Learning Research}, 21\penalty0 (140):\penalty0 1--67, 2020.

\bibitem[Schramowski et~al.(2022)Schramowski, Tauchmann, and Kersting]{schramowski2022q16}
Patrick Schramowski, Christopher Tauchmann, and Kristian Kersting.
\newblock Can machines help us answering question 16 in datasheets, and in turn reflecting on inappropriate content?
\newblock In \emph{Proceedings of the 2022 ACM Conference on Fairness, Accountability, and Transparency (FAccT)}, pages 1350--1361, 2022.
\newblock \doi{10.1145/3531146.3533192}.

\bibitem[Schramowski et~al.(2023)Schramowski, Brack, Deiseroth, and Kersting]{schramowski2023safe}
Patrick Schramowski, Manuel Brack, Bj{\"o}rn Deiseroth, and Kristian Kersting.
\newblock Safe latent diffusion: Mitigating inappropriate degeneration in diffusion models.
\newblock In \emph{Proceedings of the IEEE/CVF Conference on Computer Vision and Pattern Recognition (CVPR)}, pages 22522--22531, 2023.

\bibitem[Shazeer et~al.(2017)Shazeer, Mirhoseini, Maziarz, Davis, Le, Hinton, and Dean]{shazeer2017moe}
Noam Shazeer, Azalia Mirhoseini, Krzysztof Maziarz, Andy Davis, Quoc~V. Le, Geoffrey~E. Hinton, and Jeff Dean.
\newblock Outrageously large neural networks: The sparsely-gated mixture-of-experts layer.
\newblock In \emph{International Conference on Learning Representations (ICLR)}, 2017.

\bibitem[Shirkavand et~al.(2025)Shirkavand, Yu, Gao, Somepalli, Goldstein, and Huang]{shirkavand2025efficient}
Reza Shirkavand, Peiran Yu, Shangqian Gao, Gowthami Somepalli, Tom Goldstein, and Heng Huang.
\newblock Efficient fine-tuning and concept suppression for pruned diffusion models.
\newblock In \emph{Proceedings of the Computer Vision and Pattern Recognition Conference}, pages 18619--18629, 2025.

\bibitem[Singer et~al.(2023)Singer, Polyak, Hayes, Yin, An, Zhang, Hu, Yang, Ashual, Gafni, Parikh, Gupta, and Taigman]{singer2022make}
Uriel Singer, Adam Polyak, Thomas Hayes, Xi~Yin, Jie An, Songyang Zhang, Qiyuan Hu, Harry Yang, Oron Ashual, Oran Gafni, Devi Parikh, Sonal Gupta, and Yaniv Taigman.
\newblock {Make-A-Video}: Text-to-video generation without text-video data.
\newblock In \emph{International Conference on Learning Representations (ICLR)}, 2023.

\bibitem[Solaiman and Dennison(2021)]{solaiman2021process}
Irene Solaiman and Christy Dennison.
\newblock Process for adapting language models to society (palms) with values-targeted datasets.
\newblock \emph{Advances in Neural Information Processing Systems (NeurIPS)}, 34:\penalty0 5861--5873, 2021.

\bibitem[{Team Wan} et~al.(2025){Team Wan}, Wang, Ai, Wen, Mao, Xie, Chen, Yu, Zhao, Yang, et~al.]{wan2025wan}
{Team Wan}, Ang Wang, Baole Ai, Bin Wen, Chaojie Mao, Chen-Wei Xie, Di~Chen, Feiwu Yu, Haiming Zhao, Jianxiao Yang, et~al.
\newblock {Wan}: Open and advanced large-scale video generative models.
\newblock \emph{arXiv preprint arXiv:2503.20314}, 2025.

\bibitem[{Tencent Hunyuan Foundation Model Team}(2025)]{hunyuanvideo15}
{Tencent Hunyuan Foundation Model Team}.
\newblock {HunyuanVideo} 1.5 technical report.
\newblock \emph{arXiv preprint arXiv:2511.18870}, 2025.

\bibitem[Thakral et~al.(2025)Thakral, Glaser, Hassner, Vatsa, and Singh]{thakral2025fine}
Kartik Thakral, Tamar Glaser, Tal Hassner, Mayank Vatsa, and Richa Singh.
\newblock Fine-grained erasure in text-to-image diffusion-based foundation models.
\newblock In \emph{Proceedings of the IEEE/CVF Conference on Computer Vision and Pattern Recognition}, pages 9121--9130, 2025.

\bibitem[Tsai et~al.(2024)Tsai, Hsu, Xie, Lin, Chen, Li, Chen, Yu, and Huang]{tsai2024ringabell}
Yu-Lin Tsai, Chia-Yi Hsu, Chulin Xie, Chih-Hsun Lin, Jia-You Chen, Bo~Li, Pin-Yu Chen, Chia-Mu Yu, and Chun-Ying Huang.
\newblock {Ring-A-Bell}! how reliable are concept removal methods for diffusion models?
\newblock In \emph{International Conference on Learning Representations (ICLR)}, 2024.

\bibitem[Unterthiner et~al.(2018)Unterthiner, van Steenkiste, Kurach, Marinier, Michalski, and Gelly]{unterthiner2018fvd}
Thomas Unterthiner, Sjoerd van Steenkiste, Karol Kurach, Rapha{\"e}l Marinier, Marcin Michalski, and Sylvain Gelly.
\newblock Towards accurate generative models of video: A new metric \& challenges.
\newblock \emph{arXiv preprint arXiv:1812.01717}, 2018.

\bibitem[Wang et~al.(2023)Wang, Yuan, Chen, Zhang, Wang, and Zhang]{wang2023modelscope}
Jiuniu Wang, Hangjie Yuan, Dayou Chen, Yingya Zhang, Xiang Wang, and Shiwei Zhang.
\newblock Modelscope text-to-video technical report.
\newblock \emph{arXiv preprint arXiv:2308.06571}, 2023.

\bibitem[Weidinger et~al.(2021)Weidinger, Mellor, Rauh, Griffin, Uesato, Huang, Cheng, Glaese, Balle, Kasirzadeh, et~al.]{weidinger2021ethical}
Laura Weidinger, John Mellor, Maribeth Rauh, Conor Griffin, Jonathan Uesato, Po-Sen Huang, Myra Cheng, Mia Glaese, Borja Balle, Atoosa Kasirzadeh, et~al.
\newblock Ethical and social risks of harm from language models.
\newblock \emph{arXiv preprint arXiv:2112.04359}, 2021.

\bibitem[Wu et~al.(2025)Wu, Le, Hayat, and Harandi]{wu2025erasing}
Jing Wu, Trung Le, Munawar Hayat, and Mehrtash Harandi.
\newblock Erasing undesirable influence in diffusion models.
\newblock In \emph{Proceedings of the Computer Vision and Pattern Recognition Conference}, pages 28263--28273, 2025.

\bibitem[Wu et~al.(2024)Wu, Huang, and Wei]{wu2024mole}
Xun Wu, Shaohan Huang, and Furu Wei.
\newblock Mixture of {LoRA} experts.
\newblock In \emph{International Conference on Learning Representations (ICLR)}, 2024.

\bibitem[Xu et~al.(2024)Xu, Zhu, Zhang, Zhou, and Yu]{xu2024machine}
Heng Xu, Tianqing Zhu, Lefeng Zhang, Wanlei Zhou, and Philip~S. Yu.
\newblock Machine unlearning: A survey.
\newblock \emph{ACM Computing Surveys}, 56\penalty0 (1):\penalty0 1--36, 2024.
\newblock \doi{10.1145/3603620}.

\bibitem[Xu et~al.(2016)Xu, Mei, Yao, and Rui]{xu2016msrvtt}
Jun Xu, Tao Mei, Ting Yao, and Yong Rui.
\newblock {MSR-VTT}: A large video description dataset for bridging video and language.
\newblock In \emph{IEEE Conference on Computer Vision and Pattern Recognition (CVPR)}, pages 5288--5296, 2016.

\bibitem[Xu et~al.(2025)Xu, Zhang, Li, Chen, Zhou, Li, Du, and Ji]{xu2025videoeraser}
Naen Xu, Jinghuai Zhang, Changjiang Li, Zhi Chen, Chunyi Zhou, Qingming Li, Tianyu Du, and Shouling Ji.
\newblock Videoeraser: Concept erasure in text-to-video diffusion models.
\newblock In \emph{Proceedings of the 2025 Conference on Empirical Methods in Natural Language Processing (EMNLP)}, pages 5954--5983, 2025.

\bibitem[Yang et~al.(2024)Yang, Gao, Wang, Ho, Xu, and Xu]{yang2024mmadiffusion}
Yijun Yang, Ruiyuan Gao, Xiaosen Wang, Tsung-Yi Ho, Nan Xu, and Qiang Xu.
\newblock {MMA-Diffusion}: {MultiModal} attack on diffusion models.
\newblock In \emph{Proceedings of the IEEE/CVF Conference on Computer Vision and Pattern Recognition (CVPR)}, pages 7737--7746, 2024.

\bibitem[Yang et~al.(2025)Yang, Teng, Zheng, Ding, Huang, Xu, Yang, Hong, Zhang, Feng, Yin, Zhang, Wang, Cheng, Xu, Gu, Dong, and Tang]{yang2024cogvideox}
Zhuoyi Yang, Jiayan Teng, Wendi Zheng, Ming Ding, Shiyu Huang, Jiazheng Xu, Yuanming Yang, Wenyi Hong, Xiaohan Zhang, Guanyu Feng, Da~Yin, Yuxuan Zhang, Weihan Wang, Yean Cheng, Bin Xu, Xiaotao Gu, Yuxiao Dong, and Jie Tang.
\newblock {CogVideoX}: Text-to-video diffusion models with an expert transformer.
\newblock In \emph{International Conference on Learning Representations (ICLR)}, 2025.

\bibitem[Ye et~al.(2025)Ye, Cheng, Wang, Xiong, and Li]{ye2025t2vunlearning}
Xiaoyu Ye, Songjie Cheng, Yongtao Wang, Yajiao Xiong, and Yishen Li.
\newblock T2vunlearning: A concept erasing method for text-to-video diffusion models.
\newblock \emph{arXiv preprint arXiv:2505.17550}, 2025.

\bibitem[Yoon et~al.(2025)Yoon, Yu, Patil, Yao, and Bansal]{yoon2025safree}
Jaehong Yoon, Shoubin Yu, Vaidehi Patil, Huaxiu Yao, and Mohit Bansal.
\newblock Safree: Training-free and adaptive guard for safe text-to-image and video generation.
\newblock In \emph{International Conference on Learning Representations (ICLR)}, 2025.

\bibitem[Zhang et~al.(2024{\natexlab{a}})Zhang, Wang, Xu, Wang, and Shi]{zhang2024forget}
Gong Zhang, Kai Wang, Xingqian Xu, Zhangyang Wang, and Humphrey Shi.
\newblock {Forget-Me-Not}: Learning to forget in text-to-image diffusion models.
\newblock In \emph{Proceedings of the IEEE/CVF Conference on Computer Vision and Pattern Recognition (CVPR) Workshops}, pages 1755--1764, 2024{\natexlab{a}}.

\bibitem[Zhang et~al.(2025)Zhang, Xie, Fu, and Chen]{zhang2025concept}
Lingyun Zhang, Yu~Xie, Yanwei Fu, and Ping Chen.
\newblock Concept replacer: Replacing sensitive concepts in diffusion models via precision localization.
\newblock In \emph{Proceedings of the Computer Vision and Pattern Recognition Conference}, pages 8172--8181, 2025.

\bibitem[Zhang et~al.(2024{\natexlab{b}})Zhang, Jia, Chen, Chen, Zhang, Liu, Ding, and Liu]{zhang2024unlearndiffatk}
Yimeng Zhang, Jinghan Jia, Xin Chen, Aochuan Chen, Yihua Zhang, Jiancheng Liu, Ke~Ding, and Sijia Liu.
\newblock To generate or not? {Safety-Driven} unlearned diffusion models are still easy to generate unsafe images ... for now.
\newblock In \emph{European Conference on Computer Vision (ECCV)}, Lecture Notes in Computer Science, pages 385--403, 2024{\natexlab{b}}.
\newblock \doi{10.1007/978-3-031-72998-0_22}.

\bibitem[Zhao et~al.(2023)Zhao, Bai, Rao, Zhou, and Lu]{zhao2023unipc}
Wenliang Zhao, Lujia Bai, Yongming Rao, Jie Zhou, and Jiwen Lu.
\newblock {UniPC}: A unified predictor-corrector framework for fast sampling of diffusion models.
\newblock In \emph{Advances in Neural Information Processing Systems (NeurIPS)}, volume~36, pages 49842--49869, 2023.

\bibitem[Zheng et~al.(2024)Zheng, Peng, Yang, Shen, Li, Liu, Zhou, Li, and You]{zheng2024open}
Zangwei Zheng, Xiangyu Peng, Tianji Yang, Chenhui Shen, Shenggui Li, Hongxin Liu, Yukun Zhou, Tianyi Li, and Yang You.
\newblock Open-sora: Democratizing efficient video production for all.
\newblock \emph{arXiv preprint arXiv:2412.20404}, 2024.

\end{thebibliography}

\clearpage
\appendix
\section{Implementation details}
\label{app:impl}
This appendix collects the numerical settings and design analyses behind FADE. \cref{app:hyperparams} lists every hyperparameter used in the main paper, grouped by stage; \cref{app:cr-derivation} derives the closed-form solution for the Concept Redirect edit; \cref{app:training-algs} states the two training-stage procedures (CR and FAE) as pseudocode; \cref{app:inference-alg} does the same for the inference loop; \cref{app:edit-locus} explains why CR edits cross-attention K/V rather than the FFN; \cref{app:loss-discussion} gives the formal ESD target construction and discusses alternative loss formulations we considered; \cref{app:routing-failures} explains the routing rule; and \cref{app:composition} analyzes the composition of experts.

\section{Hyperparameters}
\label{app:hyperparams}

All three FADE stages share a common backbone configuration (Wan2.1-T2V-1.3B, $\Tlat{=}5$ for training and $\Tlat{=}21$ for inference, flow-matching UniPC $50$ steps with shift $5$ and CFG $5$ at $832{\times}480$). We instantiate the stage-specific parameters as follows.

\paragraph{Concept Redirect (\cref{sec:method:cr}).}
The closed-form K/V edit follows the UCE solver (\cref{eq:cr-argmin}) with $\lambda{=}0.5$, $s_e{=}1.0$, and $s_p{=}0.1$. The preserve set $\preserveset$ is the held-out nine Imagenette classes for object erasure, augmented by ten neutral motion phrases for style and nudity erasure.
 The Motion-Paraphrase Embedding $\embMPE_i$ averages the bare concept token with $K{=}4$ motion-paraphrased variants per concept, generated with GPT-4 and manually filtered. The Texture-Phase Decay schedule of \cref{eq:cr-blend} uses $\alpha_{\max}{=}1.0$, $\alpha_{\min}{=}0.2$, $\tsplit{=}0.4$, with $\alphaTPD$ defined as
\begin{equation}
\alphaTPD(\tnorm) \;=\;
\begin{cases}
\alpha_{\max}, & \tnorm \ge \tsplit \quad\text{(structure phase)} \\[4pt]
\alpha_{\min} + (\alpha_{\max}-\alpha_{\min})\cdot\dfrac{1 - \cos(\pi\tnorm/\tsplit)}{2}, & \tnorm < \tsplit \quad\text{(texture phase)}.
\end{cases}
\label{eq:tpd}
\end{equation}
The schedule is $C^1$-continuous at $\tnorm{=}\tsplit$ and $\tnorm{=}0$, with steepest descent at $\tnorm{=}\tsplit/2$.

\paragraph{Frame-Aware Erasure (\cref{sec:method:fae}).}
Each FA-LoRA expert uses static rank $r_0{=}4$ and frame-aware rank $r_1{=}2$, attached to the first FFN linear $\ffnone$ of every block ($\mathcal{B}{=}\{1,\dots,30\}$). FrameGate $\phic$ is a two-layer MLP with hidden width $32$ and sigmoid output, shared across the $30$ blocks of each concept. Per-concept training runs for $500$ AdamW~\citep{loshchilov2019adamw} iterations with learning rate $10^{-4}$, weight decay $10^{-4}$, in bf16, sampling timesteps from $[5, 50]$ to skip the near-clean tail. The four-term loss of \cref{eq:loss-tot} uses $\eta{=}1.0$ (ESD margin), $\alpha{=}0.5$ ($\Lpeak$ weight) with top-$k{=}3$ leakage frames, $\beta{=}0.5$ ($\Lneut$), and $\gamma{=}0.1$ ($\Lflick$).

\paragraph{C-MoE routing (\cref{sec:method:cmoe}).}
Concept reference tokens $\refvec$ are extracted once from $\textenc(\concept)$ with the EOS marker stripped. The token-level max-similarity gate of \cref{eq:cmoe-gate} has a soft cutoff at $g_{\min}{=}0.05$ below which the corresponding FA-LoRA contribution is dropped. For each concept we score its $30$ unsafe prompts and a $500$-prompt neutral motion pool augmented with $30$ unsafe prompts from every \emph{other} erased concept (cross-concept hard negatives) under \cref{eq:cmoe-gate}'s similarity function, then split into three regimes by the median gap $\Delta_c{=}\mathrm{median}(s_{\mathrm{unsafe}}){-}\mathrm{median}(s_{\mathrm{neutral}})$: when $\Delta_c{>}0.10$ we set $\routau$ at the midpoint of the two distributions and $\routT{=}\max(\Delta_c/4,0.02)$; when $0.02{<}\Delta_c{\leq}0.10$ we set $\routau$ at the neutral $90$th percentile with a fixed $\routT{=}0.05$; otherwise the expert is disabled (its gate is set to zero at inference; the table lists $\routau{=}1.0$ for this case).
 Calibrated values are listed in \cref{tab:cmoe-thresholds}.

\begin{table}[t]
  \caption{\textbf{Per-concept C-MoE routing thresholds.} Calibrated on a held-out unsafe/neutral split with cross-concept hard negatives, so that $\routau$ separates in-class from neutral and out-of-class prompts. The fallback row $\routau{=}1.000$ marks the regime where the median gap collapses (see text).}
  \label{tab:cmoe-thresholds}
  \centering
  \small
  \setlength{\tabcolsep}{7pt}
  \begin{minipage}[t]{0.48\linewidth}
    \centering
    \begin{tabular}{lcc}
      \toprule
      Imagenette object & $\routau$ & $\routT$ \\
      \midrule
      cassette player    & 0.320   & 0.050   \\
      chain saw          & 0.347   & 0.030   \\
      church             & 0.336   & 0.020   \\
      English springer   & 0.455   & 0.033   \\
      french horn        & 0.279   & 0.045   \\
      garbage truck      & 0.413   & 0.022   \\
      gas pump           & 0.324   & 0.040   \\
      golf ball          & 0.320   & 0.067   \\
      parachute          & 0.398   & 0.020   \\
      tench              & 1.000   & 0.050   \\
      \bottomrule
    \end{tabular}
  \end{minipage}\hfill
  \begin{minipage}[t]{0.48\linewidth}
    \centering
    \begin{tabular}{lcc}
      \toprule
      Painter / nudity & $\routau$ & $\routT$ \\
      \midrule
      Picasso            & 0.228   & 0.050   \\
      Van Gogh           & 0.561   & 0.049   \\
      Monet              & 0.317   & 0.050   \\
      Da Vinci           & 0.484   & 0.065   \\
      Salvador Dali      & 0.332   & 0.043   \\
      \midrule
      nudity (naked person) & 0.344 & 0.047 \\
      \bottomrule
    \end{tabular}
  \end{minipage}
\end{table}

The fallback $\routau{=}1.0$ row in \cref{tab:cmoe-thresholds} is reserved for concepts whose unsafe and neutral score distributions do not separate. \emph{tench} is the only concept in our benchmark in this regime. Separately, Wan2.1 does not reliably generate a fish from the token \texttt{tench} and renders a duck instead, so the backbone has little of the concept for FAE to suppress. Disabling the expert in this case avoids spurious modifications on out-of-class prompts that share the anchor's subword tokens.

\section{Closed-form derivation of the Concept Redirect edit}
\label{app:cr-derivation}

The closed-form edit cited in \cref{sec:method:cr} is obtained as follows. For each block $\ell$ and projection $W\!\in\!\{\xattnK,\xattnV\}$, we look for $W'$ that simultaneously redirects every erase concept $\concept_i$ to its guide anchor $\concept_i^{\star}$ and preserves the action of $W$ on every $\mathbf{p}_j\!\in\!\preserveset$ (\cref{fig:pipeline}a):
\begin{gather}
\mathcal{L}_{\mathrm{erase}}(W') \;=\; \sum_{i=1}^{\Nconcepts}\big\Vert W'\mathbf{c}_i - W\mathbf{c}_i^{\star}\big\Vert^2, \label{eq:cr-erase}\\
\mathcal{L}_{\mathrm{pres}}(W') \;=\; \sum_{j\in\preserveset}\big\Vert W'\mathbf{p}_j - W\mathbf{p}_j\big\Vert^2, \label{eq:cr-pres}\\
\Wuce^{(\ell)} \;=\; \arg\min_{W'}\,\big[s_e\,\mathcal{L}_{\mathrm{erase}}(W') + s_p\,\mathcal{L}_{\mathrm{pres}}(W') + \lambda\,\Vert W'-W\Vert_F^2\big],
\label{eq:cr-argmin}
\end{gather}
where $\mathbf{c}_i = \textemb(\textenc(\concept_i))_{[-1]}$ is the post-MLP last-content-token embedding of concept $i$ in the cross-attention context space, $\mathbf{c}_i^{\star}$ that of its guide anchor, $\mathbf{p}_j$ that of preserved phrase $j$, $s_e,s_p$ are the erase and preserve weights, and $\lambda$ regularizes the edit toward the original projection. Setting the gradient with respect to $W'$ to zero gives the closed form $\Wuce^{(\ell)} = \mathbf{A}\mathbf{B}^{-1}$ with
\begin{equation}
\mathbf{A} \;=\; \lambda\,W + s_e\!\sum_{i=1}^{\Nconcepts}(W\,\mathbf{c}_{i}^{\star})\mathbf{c}_{i}^{\!\top} + s_p\!\sum_{j\in\preserveset}(W\,\mathbf{p}_j)\mathbf{p}_j^{\!\top},\quad
\mathbf{B} \;=\; \lambda\,\mathbf{I} + s_e\!\sum_{i=1}^{\Nconcepts}\mathbf{c}_{i}\mathbf{c}_{i}^{\!\top} + s_p\!\sum_{j\in\preserveset}\mathbf{p}_j\mathbf{p}_j^{\!\top},
\label{eq:cr-AB}
\end{equation}
with the values of $s_e,s_p,\lambda$ given in \cref{app:hyperparams}. The minimisation is jointly solved over all $\Nconcepts$ erase concepts: a single matrix inversion per block per projection, applied once. \Cref{alg:cr-edit} implements this directly, with $\mathbf{c}_i$ replaced by the Motion-Paraphrase Embedding $\embMPE_i$ defined in \cref{sec:method:cr} when motion variants are available.

\section{FADE training algorithms}
\label{app:training-algs}

FADE training is a two-stage procedure. \Cref{alg:cr-edit} performs the closed-form Concept Redirect (\cref{sec:method:cr}): it produces the edited cross-attention K/V projections $\{\Wuce^{(\ell)}\}$ from the original $\{\Worig^{(\ell)}\}$ in a single linear-algebraic pass, with optional Motion-Paraphrase Embeddings (MPE, defined in \cref{sec:method:cr}) substituted for the bare-token erase anchors. The Texture-Phase Decay schedule of \cref{eq:tpd} is not part of the training stage; it is applied at inference (\cref{alg:fade}) as a per-step blend between $\Worig$ and $\Wuce$.

\Cref{alg:fae-train} trains a per-concept FA-LoRA on top of the CR-edited backbone. We follow the standard ESD teacher-student template: a frozen CR-edited base model produces three reference noise predictions per iteration (the unconditional, the unsafe-conditioned, and the neutral-conditioned), the negative-guidance target $\hat\epsilon^{\star}_{u}$ is assembled from the first two as in classical ESD, and the LoRA-active student is asked to match it on unsafe prompts and to match the frozen base on neutral prompts. The four loss terms ($\Lesd, \Lpeak, \Lneut, \Lflick$) are stated in \cref{eq:loss-tot,eq:peak-loss,eq:neut,eq:flick}; the neutral set $\mathcal{D}_{\mathrm{neu}}^{(\concept)}$ for the concept being trained pools generic motion prompts with the unsafe corpora of every other erase concept (\cref{fig:pipeline}b), which keeps the concept-specific components of different experts separated without extra data. FA-LoRA is initialized with $\loraB{0}\!=\!\mathbf{0}$, $\loraB{1}\!=\!0.01\!\cdot\!\mathrm{Kaiming}$, $\loraD{0},\loraD{1}\!=\!\mathrm{Kaiming}$, and FrameGate output bias $\mathbf{b}_2\!=\!\mathbf{0}$. The per-token frame index $\fnorm$ is read off the patchification grid as described in \cref{sec:method:fae}.

\clearpage
\begin{algorithm}[H]
\caption{Concept Redirect closed-form K/V edit (Phase 1).}
\label{alg:cr-edit}
\begin{algorithmic}[1]
\Require frozen Wan2.1 backbone $f_\theta$ exposing $\textemb\!\circ\!\textenc$ and $\{\Worig^{(\ell)}\}$ at every $\ell\!=\!1,\dots,L$ for both $\xattnK$ and $\xattnV$; erase concepts $\cset_e=\{\concept_1,\dots,\concept_{\Nconcepts}\}$ with optional motion variants $\motionset(\concept_i)$; per-concept guide anchors $\{\concept_i^{\star}\}$; preserve set $\preserveset$; scalars $\lambda$, erase scale $s_e$, preserve scale $s_p$.
\State \textbf{Anchor encoding.} For every $s\in\cset_e\cup\{\concept_i^{\star}\}\cup\preserveset\cup\bigcup_i\motionset(\concept_i)$, set $\mathbf{c}_s\gets\textemb(\textenc(s))_{[-1]}$. \Comment{post-MLP last-content-token}
\State \textbf{Motion-Paraphrase Embedding.}
\For{$\concept_i\in\cset_e$ with $\motionset(\concept_i)\ne\emptyset$}
  \State $\mathbf{c}_{\concept_i}\gets\embMPE_i = \tfrac{1}{|\motionset(\concept_i)|+1}\!\big(\mathbf{c}_{\concept_i} + \sum_{m\in\motionset(\concept_i)}\!\mathbf{c}_m\big)$ \Comment{Motion-Paraphrase Embedding}
\EndFor
\For{$\ell=1,\dots,L$ \textbf{and} $W\in\{\xattnK,\xattnV\}$}
  \State $\mathbf{A}\gets \lambda\,W + s_e\!\sum_{i=1}^{\Nconcepts}(W\,\mathbf{c}_{\concept_i^{\star}})\mathbf{c}_{\concept_i}^{\!\top} + s_p\!\sum_{j\in\preserveset}(W\,\mathbf{c}_j)\mathbf{c}_j^{\!\top}$
  \State $\mathbf{B}\gets \lambda\,\mathbf{I} + s_e\!\sum_{i=1}^{\Nconcepts}\mathbf{c}_{\concept_i}\mathbf{c}_{\concept_i}^{\!\top} + s_p\!\sum_{j\in\preserveset}\mathbf{c}_j\mathbf{c}_j^{\!\top}$
  \State Replace $W$ with $\mathbf{A}\mathbf{B}^{-1}$ in-place. \Comment{closed-form solution of \cref{eq:cr-argmin}}
\EndFor
\State \Return $\{\Wuce^{(\ell)}\}_{\ell=1}^L$ for both K and V projections
\end{algorithmic}
\end{algorithm}

\begin{algorithm}[H]
\caption{Frame-Aware Erasure training, per concept $\concept$ (Phase 2).}
\label{alg:fae-train}
\begin{algorithmic}[1]
\Require CR-edited base $f_\theta^{(\mathrm{CR})}$ with $\{\Wuce^{(\ell)}\}$ frozen; concept $\concept$; unsafe corpus $\mathcal{D}_{\mathrm{u}}^{(\concept)}$; neutral set $\mathcal{D}_{\mathrm{neu}}^{(\concept)}=\mathcal{D}_{\mathrm{neu\text{-}motion}}\cup\bigcup_{\concept'\ne\concept}\mathcal{D}_{\mathrm{u}}^{(\concept')}$; LoRA ranks $r_0,r_1$; loss weights $\eta,\alpha,\beta,\gamma$; top-$k$; learning rate $\mathrm{lr}$; iterations $N$; training-step range $[t_{\min},t_{\max}]$.
\State Attach per-block FA-LoRA $\Wlora^{(\ell)}_{\concept}=(\loraB{0},\loraD{0},\loraB{1},\loraD{1},\phic)$ for $\ell\!\in\!\mathcal{B}$ with $\loraB{0}\!=\!\mathbf{0}$, $\loraB{1}\!=\!0.01\!\cdot\!\mathrm{Kaiming}$, $\loraD{0},\loraD{1}\!=\!\mathrm{Kaiming}$, FrameGate bias $\mathbf{b}_2\!=\!\mathbf{0}$.
\State Pre-encode contexts $\textemb\!\circ\!\textenc$ for $\mathcal{D}_{\mathrm{u}}^{(\concept)}$, $\mathcal{D}_{\mathrm{neu}}^{(\concept)}$, and the empty prompt.
\For{$i=1,\dots,N$}
  \State sample $p^{-}\!\sim\!\mathcal{D}_{\mathrm{u}}^{(\concept)}$,\;\, $p^{\mathrm{neu}}\!\sim\!\mathcal{D}_{\mathrm{neu}}^{(\concept)}$,\;\, $t\!\sim\!\mathrm{Uniform}[t_{\min},t_{\max}]$,\;\, seed.
  \State $\mathbf{x}_t\gets$ partial-denoise $f_\theta^{(\mathrm{CR})}$ on $p^{-}$ to step $t$ with classifier-free guidance. \Comment{base only, no grad}
  \State \textbf{Three frozen forwards on $f_\theta^{(\mathrm{CR})}$ at $(\mathbf{x}_t,t)$:}
  \State \quad $\hat\epsilon_{\varnothing}\gets f_\theta^{(\mathrm{CR})}(\mathbf{x}_t,t,\varnothing);\;\;\hat\epsilon_{u}\gets f_\theta^{(\mathrm{CR})}(\mathbf{x}_t,t,p^{-});\;\;\hat\epsilon^{0}_{n}\gets f_\theta^{(\mathrm{CR})}(\mathbf{x}_t,t,p^{\mathrm{neu}})$
  \State $\hat\epsilon^{\star}_{u}\gets\hat\epsilon_{\varnothing}-\eta(\hat\epsilon_{u}-\hat\epsilon_{\varnothing})$ \Comment{ESD negative-guidance target}
  \State \textbf{Two LoRA-active forwards on $f_\theta^{(\mathrm{CR})}\!+\!\Wlora_{\concept}$ with concept gate $\gate\!=\!1$:}
  \State \quad $\hat\epsilon^{\mathrm{LoRA}}_{u}\gets f_\theta^{(\mathrm{CR},\Wlora)}(\mathbf{x}_t,t,p^{-});\;\;\hat\epsilon^{\mathrm{LoRA}}_{n}\gets f_\theta^{(\mathrm{CR},\Wlora)}(\mathbf{x}_t,t,p^{\mathrm{neu}})$
  \State $\Lesd\gets\big\Vert\hat\epsilon^{\mathrm{LoRA}}_{u}-\hat\epsilon^{\star}_{u}\big\Vert_2^2$
  \State $\Lpeak\gets\tfrac{1}{k}\!\sum_{f\in\mathrm{topk}_k(\delta_f)}\!\delta_f,\quad \delta_f=\tfrac{1}{|S_f|}\Vert\hat\epsilon^{\mathrm{LoRA}}_{u}[f]-\hat\epsilon^{\star}_{u}[f]\Vert_2^2$ \Comment{\cref{eq:peak-loss}}
  \State $\Lneut\gets\big\Vert\hat\epsilon^{\mathrm{LoRA}}_{n}-\hat\epsilon^{0}_{n}\big\Vert_2^2$ \Comment{\cref{eq:neut}, near-identity on neutral set}
  \State $\Lflick\gets\tfrac{1}{\Tlat-1}\!\sum_{f=1}^{\Tlat-1}\!\big\Vert\phic(\fnorm_{f+1},\tnorm)-\phic(\fnorm_f,\tnorm)\big\Vert_2^2$ at the shallowest $\ell\!\in\!\mathcal{B}$ \Comment{\cref{eq:flick}}
  \State $\Ltot\gets\Lesd+\alpha\Lpeak+\beta\Lneut+\gamma\Lflick$ \Comment{\cref{eq:loss-tot}}
  \State AdamW step on $\{\loraB{0},\loraD{0},\loraB{1},\loraD{1},\phic\}_{\ell\in\mathcal{B}}$ with learning rate $\mathrm{lr}$.
\EndFor
\State \Return per-concept FA-LoRA bundle $\Wlora_{\concept}=\{\Wlora^{(\ell)}_{\concept}\}_{\ell\in\mathcal{B}}$.
\end{algorithmic}
\end{algorithm}
\clearpage

\section{FADE inference algorithm}
\label{app:inference-alg}

\Cref{alg:fade} states the full inference procedure of FADE on a single prompt, fusing CR, FAE, and C-MoE into one denoising loop. The C-MoE gates are computed once at the encoder level; TPD reblends the cross-attention K/V projections at the start of every denoising step; and the active FAE-experts are summed into the FFN path at every block within $\mathcal{B}$, frame-by-frame. When the active set $\mathcal{A}$ is empty, as for most clean prompts, the procedure reduces to CR alone.

\begin{algorithm}[H]
\caption{FADE inference on prompt $p$.}
\label{alg:fade}
\begin{algorithmic}[1]
\Require prompt $p$; CR-edited projections $\{\Wuce^{(\ell)}\}_\ell$ alongside the originals $\{\Worig^{(\ell)}\}_\ell$; per-concept $\{\Wlora_\concept,\phic,\refvec,\routau,\routT\}_{\concept\in\cset_e}$; gate cutoff $g_{\min}$; denoising schedule $\{t_n\}$
\State compute $\gate \gets \sigma\!\big((\mathrm{sim}(p,\concept)-\routau)/\routT\big)$ for every enabled $\concept\in\cset_e$; set $\gate\gets 0$ for disabled experts \Comment{\cref{eq:cmoe-gate}}
\State active set $\mathcal{A}\gets\{\concept : \gate \ge g_{\min}\}$ \Comment{usually empty for neutral prompts}
\State sample $\mathbf{x}_{t_N}\sim\mathcal{N}(\mathbf{0},\mathbf{I})$
\For{$n = N,\dots,1$}
  \State $\tnorm \gets t_n / t_N$;\;\; $\alphaTPD\gets$ schedule of \cref{eq:tpd}
  \For{each block $\ell$}
    \State set cross-attn K/V to $(1-\alphaTPD)\Worig^{(\ell)} + \alphaTPD\,\Wuce^{(\ell)}$ \Comment{TPD blend, \cref{eq:cr-blend}}
  \EndFor
  \For{each block $\ell\in\mathcal{B}$}
    \State $\Delta_{\ell}(\fnorm,\tnorm)\gets \sum_{\concept\in\mathcal{A}} \gate\,\Wlora^{(\ell)}_{\concept}(\fnorm,\tnorm)$ \Comment{additive FAE-expert mixture}
    \State replace FFN[0] forward by $\mathbf{x}\mapsto \ffnone\mathbf{x} + \Delta_{\ell}(\fnorm,\tnorm)\mathbf{x}$
  \EndFor
  \State $\hat\epsilon_{t_n}\gets f_\theta(\mathbf{x}_{t_n}, p)$;\;\; $\mathbf{x}_{t_{n-1}}\gets\mathrm{step}(\mathbf{x}_{t_n},\hat\epsilon_{t_n}, t_n)$
\EndFor
\State \Return decoded video from $\mathbf{x}_{t_0}$
\end{algorithmic}
\end{algorithm}

\section{Architectural rationale: why CR edits cross-attention K/V, not the FFN}
\label{app:edit-locus}

We edit the cross-attention K/V projections because they are the direct interface to text conditioning in Wan2.1, whereas several T2I and earlier T2V unlearning methods (notably the closed-form refusal-vector edit of~\citet{facchiano2025video}) target the FFN. The observations below come from pilot experiments and are not quantified in the paper.

\paragraph{Where text enters the transformer.} The closed-form refusal-vector edit of~\citet{facchiano2025video} was developed for a dual-stream MMDiT (Open-Sora 2.0), in which each block has separate text and video FFNs. Editing the text FFN acts on text-only activations, in the same hidden space as the refusal vector extracted there. The Wan2.1 blocks process only video tokens: each block runs self-attention over the video tokens, cross-attention to the text context supplied as keys and values, and a single FFN. There is no text FFN. The single FFN consumes a mixture of self-attention and cross-attention residuals, so a closed-form edit of it (a) acts on activations dominated by video content and (b) perturbs spatio-temporal modeling on every prompt, including neutral ones. In pilot experiments this produced over-sharpened static textures and damaged motion on neutral prompts.

\paragraph{Cross-attention K/V as the text interface.} In Wan2.1 the only path through which text content can influence the generation is the cross-attention K/V projection of each block; the post-text-embedding context $\textemb(\textenc(p))$ is consumed only there. Editing $\xattnK$ and $\xattnV$ therefore acts where text enters the video stream and leaves the FFN unchanged by CR. The K/V projections are linear maps from the post-MLP context space, where the concept references $\embMPE_i\in\mathbb{R}^{\dimm}$ also live, so the edit is a ridge regression in that space; $\lambda>0$ makes its normal matrix invertible.

\paragraph{Why concept references are in post-MLP space, not in raw T5 space.} A natural alternative is to extract the refusal direction in the raw T5 hidden space ($\dimT{=}4096$) and project it into the DiT context space when applying the edit. This fails because $\textemb$ contains a GELU non-linearity, so directions in T5 space do not map linearly to directions in context space; the closed-form edit equation becomes ill-defined. Sampling concept references at the post-MLP last-content-token position is the smallest change that restores linearity and matches the K/V projections' input space.

\paragraph{Alternative edit sites and their costs.} Three other sites were considered before settling on cross-attention K/V. Editing the video-side query $W^{(\ell)}_Q$ corrupts every cross-attention head on every prompt, including neutral ones, because the query is not concept-specific. Editing the FFN's first linear $\ffnone$ matches the locus used by~\citet{facchiano2025video}, but its input is mixed, and in pilot runs the same closed-form solver applied at $\ffnone$ degraded motion on neutral prompts. Editing self-attention projections is similarly text-blind. None of the three isolates text conditioning as directly as K/V editing, so CR edits K/V only and FAE addresses the residual frame-reactivation gap through a separate, parametric correction on the FFN path.

\section{Loss-design discussion}
\label{app:loss-discussion}

We first give the formal construction of the ESD-style target $y_t$ that defines $\Lesd$ in \cref{eq:loss-tot}, then describe three design choices made on the basis of pilot runs (not quantified in the paper).

\paragraph{ESD target construction.} Let $\hat\epsilon^{0}_t$ denote the noise prediction of the frozen CR model at timestep $t$. For an unsafe prompt $p^{-}$ and margin $\eta\!>\!0$, the negative-guidance target $y_t$ and the loss are
\begin{equation}
y_t \;=\; \hat\epsilon^{0}_t(\mathbf{x}_t,\varnothing) - \eta\,\big(\hat\epsilon^{0}_t(\mathbf{x}_t,p^{-}) - \hat\epsilon^{0}_t(\mathbf{x}_t,\varnothing)\big),\qquad
\Lesd \;=\; \big\Vert\hat\epsilon_t(\mathbf{x}_t,p^{-}) - y_t\big\Vert_2^2,
\label{eq:esd}
\end{equation}
with $\varnothing$ the unconditional context. This is the negative-guidance objective of ESD~\citep{gandikota2023erasing} with the CR model as teacher, so the FA-LoRA only has to erase the residual concept signal surviving CR. The construction is fully self-supervised: the frozen teacher synthesizes $y_t$ for every sampled $(\mathbf{x}_t,t,p^{-})$, removing the need for ``concept-free'' ground-truth videos that would be expensive and impossible to define cleanly. Because $y_t$ is built from the frozen teacher's own predictions, the student is supervised in the same prediction space as the backbone.

\paragraph{Noise-space versus feature-space supervision for $\Lneut$.} A natural alternative to enforcing identity at the predicted-noise level is to enforce identity at an intermediate FFN feature, e.g.\ matching the FFN output of the LoRA-on and LoRA-off branches at every block of $\mathcal{B}$. In pilot runs this produced over-regularized LoRAs: the static rank $r_0$ branch is locked toward zero across all blocks, so the LoRA can no longer suppress the target concept either. Operating $\Lneut$ in the noise space lets the LoRA freely re-route through earlier blocks as long as the final prediction on neutral prompts matches the frozen CR model, which preserves enough capacity for the concept-suppression signal.

\paragraph{Top-$k$ rather than mean for $\Lpeak$.} The peak-frame leakage term penalizes the worst $k$ frames per clip, not the mean over all frames. In pilot runs we compared $k\!=\!\Tlat$ (mean), $k\!=\!1$ (single argmax), and $k\!=\!3$. The mean variant gives smoother gradients but did not close $\reactgap$, because it permits the model to trade clean middle frames against a single high-leakage spike. With $k\!=\!1$, gradients alternate between different worst frames on nearby steps, which made optimization unstable. With $k\!=\!3$, training was stable and still captured leakage spikes, so we use it throughout.

\paragraph{Evaluating $\Lflick$ once.} All blocks of a concept share one FrameGate, so its output is identical at every depth, and we evaluate the smoothness penalty once per training example. Applying it once per block would multiply its weight by the number of blocks; in pilot runs this over-smoothed the gate toward a constant. The gate receives its frame-dependent signal through $\Lpeak$, i.e., through the per-frame residuals of the prediction.

\section{Routing design rationale}
\label{app:routing-failures}

We motivate the token-level max-similarity rule of \cref{eq:cmoe-gate} by reporting how a pooled-cosine router separated the concepts we tested.

\paragraph{Pooled cosine routing fails on motion-in-scene concepts.} A pooled-cosine baseline scores each prompt by a single cosine between a pooled prompt embedding and a pooled concept anchor. Across four pooling-and-space combinations (mean-of-tokens vs.\ last-content-token, post-MLP context space vs.\ raw T5 space, with and without global-mean centering), unsafe-vs-neutral median margins on five test concepts (\emph{english springer spaniel}, \emph{parachute}, \emph{church}, \emph{Van Gogh}, \emph{anime}) sit within $|\Delta|\!<\!0.025$ in cosine units, below what a sigmoid gate at $T{=}0.05$ can resolve. Centering helps on object-free concepts but \emph{hurts} on \emph{spaniel} and \emph{church}: their unsafe prompts share scene statistics (outdoor, motion) with the neutral pool, so the global mean strips the very component carrying signal. Pooling that mixes the concept token with scene tokens separated object-in-scene concepts poorly.

\paragraph{Token-level max-similarity recovers the concept token.} Replacing the pooled cosine with $\max_{i,j}\cos(\textenc(p)_i,\,\mathbf{r}_{\concept,j})$ over all (prompt token, concept-anchor token) pairs lifts every test concept's median margin above $+0.04$, resolving the under-separation. The rule trades an aggregated representation for the single most concept-aligned token pair, which is precisely what pooling discarded. Centering, useful as a tie-breaker under pooling, is harmful here because it rotates the concept tokens off-axis without adding separation.

\paragraph{Why we keep multi-token anchors.} A natural shortening is to use a discriminative substring of each multi-word concept (e.g.\ \emph{spaniel} for \emph{english springer spaniel}) to avoid token-level false positives such as \emph{``An English garden''}. Numerically, shortened anchors keep all margins above $0.02$, but we do not adopt them for two reasons: (a) the bare anchor for \emph{spaniel} matches sibling breeds (\emph{cocker}, \emph{water}, \emph{field}) the FA-LoRA was not trained to handle, producing collateral damage on out-of-target spaniels; (b) the routing scope must match the user contract, so ``unlearn english springer spaniel'' should not silently fire on every spaniel breed. We accept the small false-positive rate from modifier tokens (e.g.\ \emph{english}, \emph{van}) as the cost of preserving named-concept semantics.

\section{Why FAE experts compose with little interference}
\label{app:composition}

Adding $\Nconcepts$ independently trained FA-LoRAs at the same FFN linear is not safe in general: inconsistent updates on a shared token corrupt the FFN output. The additive composition used in \cref{sec:method:cmoe} relies on three mechanisms; \cref{app:geometry} measures the resulting geometry.

\paragraph{(i) Functional near-identity via $\Lneut$.} Every FA-LoRA is trained to leave the noise prediction unchanged outside its own concept: $\Lneut$ (\cref{eq:neut}) includes the unsafe prompts of every other erased concept (\cref{fig:pipeline}b), so expert $\concept_a$ has little effect on the prediction for a prompt of concept $\concept_b$. This is a constraint on behavior, not on weights: the measured updates share a common suppression template and differ in well-separated concept-specific components (\cref{app:geometry}). The separation comes from the training data rather than from an explicit subspace constraint.

\paragraph{(ii) Small perturbations (assumption).} Low rank ($r_0,r_1$ in \cref{app:hyperparams}), a zero-initialized static branch, a small-$\loraB{1}$ initialization, and a sigmoid-bounded FrameGate keep each $\Wlora_{\concept}$ small at the start of training and bound the gate; we assume the trained updates remain in the regime where the network responds approximately linearly to them. This is an assumption, not a guarantee.

\paragraph{(iii) Gate pruning.} The gate cutoff and cross-concept hard-negative calibration of \cref{eq:cmoe-gate} (described in \cref{sec:method:cmoe} and \cref{app:routing-failures}) keep the active set small: typically one or two experts fire on a concept-mentioning prompt and usually none on a clean prompt, although shared modifier tokens can cause false activations. Compositional prompts that mention several erased concepts at once---the regime where hard top-1 routing fails---are handled by firing the corresponding experts together.

\paragraph{Cross-talk bound from $\Lneut$ convergence.}
We now derive a leading-order bound that ties the empirical $\Lneut$ training residual to the composition cross-talk, making (i)--(iii) above quantitative. Let $\hat\epsilon^{0}_t(p)$ denote the frozen base (CR-edited, no FAE) noise prediction, and let $\hat\epsilon_t(p;\mathcal{A},\{g_i\})$ denote the prediction when FA-LoRAs $\{\Wlora_{\concept_i}\}_{i\in\mathcal{A}}$ are additively composed at the FFN path with gate magnitudes $\{g_i\}\!\subseteq\![0,1]$. The gates $g_i=g_i(p)$ depend on the prompt and are treated as random variables below. Define the cross-concept noise residual
\[
\mathcal{R}_{i\to j}\;\triangleq\;\mathbb{E}_{p\sim\mathcal{D}_{\mathrm{u}}^{(\concept_j)},\,\mathbf{x}_t,\,t}\,\big\Vert\hat\epsilon_t(p;\{\concept_i\},g_i{=}1)-\hat\epsilon^{0}_t(p)\big\Vert_2^2,
\]
the perturbation that LoRA-$\concept_i$ alone injects into the noise prediction on $\concept_j$'s unsafe distribution. Because $\mathcal{D}_{\mathrm{neu}}^{(\concept_i)}\supseteq\mathcal{D}_{\mathrm{u}}^{(\concept_j)}$ for every $j\!\neq\!i$ by construction (\cref{fig:pipeline}b), each $\mathcal{R}_{i\to j}$ is a component of the $\Lneut$ loss for LoRA-$\concept_i$ and is jointly minimized at training convergence.

\emph{Linearization assumption.} Under assumption (ii), the LoRA-active prediction is to first order additive across $\mathcal{A}$ and linear in each gate $g_i$:
\begin{equation}
\hat\epsilon_t(p;\mathcal{A},\{g_i\})\;\approx\;\hat\epsilon^{0}_t(p)+\sum_{i\in\mathcal{A}}g_i\,\big[\hat\epsilon_t(p;\{\concept_i\},g_i{=}1)-\hat\epsilon^{0}_t(p)\big].
\label{eq:lora-linearization}
\end{equation}
This is the standard Taylor expansion of the network in the parameter-perturbation $\sum_i g_i\Wlora_{\concept_i}$ around the base point.

For any prompt $p\!\in\!\mathcal{D}_{\mathrm{u}}^{(\concept_j)}$ with $j\!\in\!\mathcal{A}$, applying \cref{eq:lora-linearization} to both forward passes and subtracting cancels the $j$-th term, leaving the off-target sum $\sum_{i\in\mathcal{A}\setminus\{j\}}g_i\,[\hat\epsilon_t(p;\{\concept_i\},g_i{=}1)-\hat\epsilon^{0}_t(p)]$. Taking $\mathbb{E}_{p,\mathbf{x}_t,t}\Vert\!\cdot\!\Vert_2$ on both sides and applying the triangle inequality and then the Cauchy--Schwarz inequality to each term yields, to leading order,
\begin{equation}
\mathbb{E}\big\Vert\hat\epsilon_t(p;\mathcal{A},\{g_i\})-\hat\epsilon_t(p;\{\concept_j\},g_j)\big\Vert_2
\;\lesssim\;\sum_{i\neq j}\sqrt{\mathbb{E}\big[g_i^2\big]\,\mathcal{R}_{i\to j}},
\label{eq:cross-talk-bound}
\end{equation}
where $g_i{=}0$ for experts outside $\mathcal{A}$ and the expectation is over $p\sim\mathcal{D}_{\mathrm{u}}^{(\concept_j)}$. The bound has three properties that mirror mechanisms (i)--(iii). It is linear in each training residual $\sqrt{\mathcal{R}_{i\to j}}$ that the cross-concept hard-negative supervision of $\Lneut$ drives toward zero (mechanism (i)). It scales with the root-mean-square gate of each off-target expert on $\concept_j$'s prompts, which the cutoff $g_{\min}$ and the calibration keep small (mechanism (iii)); the linearization in \cref{eq:lora-linearization} rests on assumption (ii). To leading order, cross-talk under composition is therefore controlled by quantities that $\Lneut$ and the gate calibration jointly minimize. The bound is a first-order argument, not a guarantee, and it does not cover prompts that contain both concepts; \cref{tab:multi}(a) evaluates such prompts empirically.

\paragraph{Residual failure regime.} Mechanisms (i)--(iii) leave one regime unaddressed: directly opposed semantics on overlapping inputs (e.g.\ ``add skin texture'' vs.\ ``remove skin texture''), where two experts would write conflicting updates to the same token. No pair of concepts in our benchmarks requires opposed updates, so this regime lies outside our experiments; we leave its treatment to future work.

\section{Additional quantitative results}
\label{app:more-results}

\subsection{Unsafe categories and celebrity identities}
\label{app:unsafe-celeb}
\paragraph{Unsafe content.}
We follow the nudity protocol of ESD~\citep{gandikota2023erasing}, UCE~\citep{gandikota2024unified}, and MACE~\citep{lu2024mace}, adopted for video by VideoEraser and Refusal: we generate the I2P prompts~\citep{schramowski2023safe} (4{,}703 prompts, 5 seeds each) and check every frame with NudeNet~\citep{praneeth2019nudenet}. We add the violence and hate categories of I2P, judged by Q16~\citep{schramowski2022q16}, the standard judge for non-sexual I2P categories. \Cref{tab:unsafe-celeb}(a) reports the percentage of flagged frames: FADE lowers it from 84\% to 19\% for nudity and to 16--17\% for violence and hate, less than half of VideoEraser's rate in each category. 

\paragraph{Celebrity identities.}
Adapting the celebrity evaluation of MACE~\citep{lu2024mace}, we erase 30 celebrity identities simultaneously from one Wan2.1-1.3B model, keep 10 other identities as a retain set, and score every frame with the GIPHY Celebrity Detector~\citep{giphy2019gcd}. \Cref{tab:unsafe-celeb}(b) reports the top-1 recognition rate on erased ($\mathrm{Acc}_e$, efficacy) and retained ($\mathrm{Acc}_s$, specificity) identities. FADE reaches $\mathrm{Acc}_e{=}6.4$ while keeping $\mathrm{Acc}_s{=}86.1$ (unedited 90.8); VideoEraser erases less ($19.8$) and loses 16.3 points of retain accuracy. This is also our largest concept set, $\Nconcepts{=}30$.

\begin{table}[h]
  \caption{(a) \textbf{Unsafe content on I2P}, frames flagged by NudeNet (nudity) or Q16 (violence, hate). (b) \textbf{30 erased celebrities}: GIPHY detector accuracy on erased ($\mathrm{Acc}_e$) and retained ($\mathrm{Acc}_s$) identities.}
  \label{tab:unsafe-celeb}
  \centering
  \footnotesize
  \begin{subtable}[t]{0.50\linewidth}
    \centering
    \caption{Flagged frames (\%) $\downarrow$}
    \setlength{\tabcolsep}{4pt}
    \begin{tabular}{lcccc}
      \toprule
      Method & Nudity & Violence & Hate & Avg \\
      \midrule
      Wan2.1       & 84 & 68 & 73 & 75.0 \\
      \midrule
      Refusal      & 54 & 46 & 47 & 49.0 \\
      VideoEraser  & 42 & 35 & 38 & 38.3 \\
      \ours \textbf{FADE} & \textbf{19} & \textbf{16} & \textbf{17} & \textbf{17.3} \\
      \bottomrule
    \end{tabular}
  \end{subtable}\hfill
  \begin{subtable}[t]{0.47\linewidth}
    \centering
    \caption{30-identity erasure}
    \setlength{\tabcolsep}{4pt}
    \begin{tabular}{lccc}
      \toprule
      Method & $\mathrm{Acc}_e\!\downarrow$ & $\mathrm{Acc}_s\!\uparrow$ & FVD$\downarrow$ \\
      \midrule
      Wan2.1       & 88.6 & 90.8 & 147.3 \\
      \midrule
      Refusal      & 60.3 & 71.2 & 189.7 \\
      VideoEraser  & 19.8 & 74.5 & 209.4 \\
      \ours \textbf{FADE} & \textbf{6.4} & \textbf{86.1} & \textbf{162.8} \\
      \bottomrule
    \end{tabular}
  \end{subtable}
\end{table}

\subsection{Per-class results on other backbones}
\label{app:backbone-tables}
\begin{figure}[h]
  \centering
  \begin{minipage}[t]{0.32\linewidth}
    \centering
    \includegraphics[width=\linewidth]{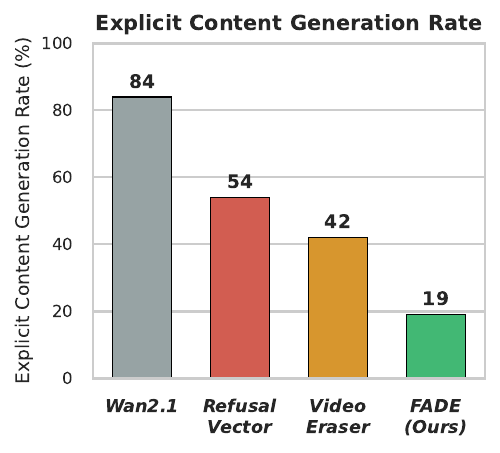}\\[-2pt]
    {\small (a) Wan2.1-T2V-1.3B}
  \end{minipage}\hfill
  \begin{minipage}[t]{0.32\linewidth}
    \centering
    \includegraphics[width=\linewidth]{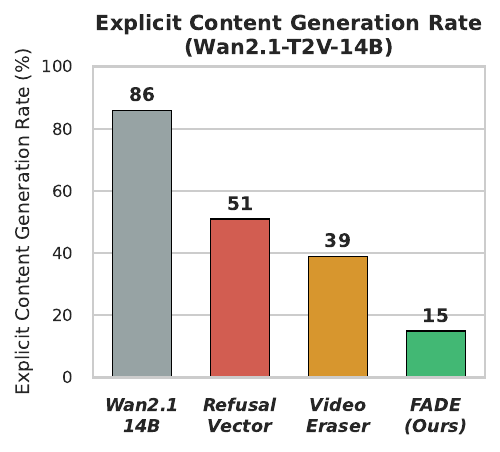}\\[-2pt]
    {\small (b) Wan2.1-T2V-14B}
  \end{minipage}\hfill
  \begin{minipage}[t]{0.32\linewidth}
    \centering
    \includegraphics[width=\linewidth]{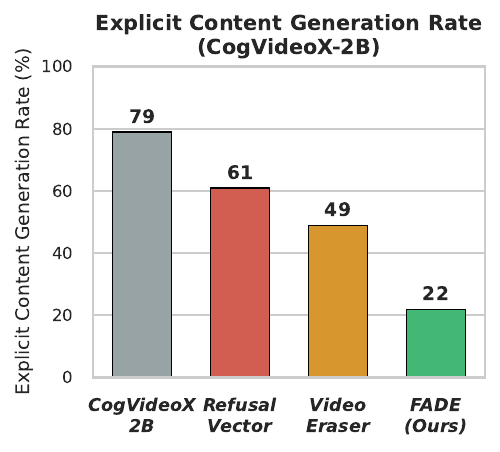}\\[-2pt]
    {\small (c) CogVideoX-2B}
  \end{minipage}
  \caption{\textbf{Nudity erasure on I2P across three backbones.} Frames flagged by NudeNet (\%).}
  \label{fig:nudity-backbones}
  \label{fig:nudity-bar}
\end{figure}

\Cref{tab:obj-cogvideox,tab:obj-main-14b} give the per-class results behind \cref{tab:backbones} for the five Imagenette classes evaluated on every backbone. Each FADE model erases all Imagenette classes at once, and the average is taken over the five listed classes. \Cref{fig:nudity-backbones} gives the corresponding nudity results.

\begin{table}[h]
  \caption{\textbf{Object erasure on CogVideoX-2B.} Per-class $\meanframeacc$ (\%) $\downarrow$ over 5 prompts $\times$ 5 seeds per class, their average, and FVD on the neutral MSR-VTT pool.}
  \label{tab:obj-cogvideox}
  \centering
  \small
  \setlength{\tabcolsep}{8pt}
  \begin{tabular}{lcccc}
    \toprule
    Class & CogVideoX-2B & Refusal & VideoEraser & \cellcolor{oursrow}\textbf{FADE} \\
    \midrule
    Chain Saw          & 68.5 & 51.3 & 21.7 & \cellcolor{oursrow}\textbf{6.8} \\
    Church             & 75.1 & 55.6 & 19.4 & \cellcolor{oursrow}\textbf{5.5} \\
    French Horn        & 93.7 & 68.9 & 20.3 & \cellcolor{oursrow}\textbf{6.7} \\
    Garbage Truck      & 74.6 & 57.2 & 13.8 & \cellcolor{oursrow}\textbf{5.3} \\
    Parachute          & 95.8 & 74.1 & 16.5 & \cellcolor{oursrow}\textbf{7.2} \\
    \midrule
    Avg $\meanframeacc\!\downarrow$ & 81.5 & 61.4 & 18.3 & \cellcolor{oursrow}\textbf{6.3} \\
    FVD$\downarrow$    & 152.6 & 191.3 & 215.7 & \cellcolor{oursrow}\textbf{165.4} \\
    \bottomrule
  \end{tabular}
\end{table}

\begin{table}[h]
  \caption{\textbf{Object erasure on Wan2.1-T2V-14B.} Same protocol as \cref{tab:obj-cogvideox}.}
  \label{tab:obj-main-14b}
  \centering
  \small
  \setlength{\tabcolsep}{8pt}
  \begin{tabular}{lcccc}
    \toprule
    Class & Wan2.1-14B & Refusal & VideoEraser & \cellcolor{oursrow}\textbf{FADE} \\
    \midrule
    Chain Saw          & 82.5 & 63.7 & 24.6 & \cellcolor{oursrow}\textbf{7.8} \\
    Church             & 86.3 & 67.4 & 22.1 & \cellcolor{oursrow}\textbf{6.2} \\
    French Horn        & 98.6 & 79.5 & 23.8 & \cellcolor{oursrow}\textbf{7.5} \\
    Garbage Truck      & 85.1 & 68.3 & 16.4 & \cellcolor{oursrow}\textbf{6.0} \\
    Parachute          & 99.1 & 84.2 & 19.7 & \cellcolor{oursrow}\textbf{8.3} \\
    \midrule
    Avg $\meanframeacc\!\downarrow$ & 90.3 & 72.6 & 21.3 & \cellcolor{oursrow}\textbf{7.2} \\
    FVD$\downarrow$    & 118.5 & 162.7 & 189.3 & \cellcolor{oursrow}\textbf{134.6} \\
    \bottomrule
  \end{tabular}
\end{table}

\subsection{Porting FADE to other backbones}
\label{app:backbones}
FADE needs two interfaces in each transformer block: the linear maps that turn text tokens into attention keys and values, which CR edits, and the first FFN linear on the video path, which carries the FA-LoRA experts. The three backbone families expose them differently. Wan2.1~\citep{wan2025wan} injects umT5~\citep{chung2023umt5} text embeddings through cross-attention, so CR edits the cross-attention K/V projections. HunyuanVideo-1.5~\citep{hunyuanvideo15} is built from 54 dual-stream blocks in which text and video tokens have separate QKV and MLP weights and meet in a joint attention; CR edits the text-stream K/V projections and FA-LoRA attaches to the video-stream MLP. CogVideoX~\citep{yang2024cogvideox} concatenates T5~\citep{raffel2020t5} text tokens with video tokens and applies the same QKV and FFN weights to both, with modality-specific adaptive LayerNorm; the closed-form solve is computed on text-token inputs as for the other backbones, but the edited projection is shared with the video tokens.

All FAE hyperparameters (loss weights $\eta,\alpha,\beta,\gamma$, ranks $r_0,r_1$, FrameGate width, 500 iterations) are the same on every backbone; only the router thresholds $(\routau,\routT)$ are recalibrated by the automatic procedure of \cref{app:hyperparams}, without manual tuning.

\subsection{VLM judge and human study}
\label{app:human-study}
\paragraph{VLM judge.} We query Qwen3.6-27B~\citep{qwen36} on every decoded frame at temperature 0 and ask for a structured answer stating whether the target concept is present.
 The per-frame answers replace $s_V[f]$ in $\meanframeacc$; everything else follows \cref{sec:exp:object}.

\paragraph{Human study.} Clips generated by the four systems in \cref{tab:t2i-judges}(b) on the same prompt-seed grid were pooled and shuffled, with the method identity hidden from annotators. Several annotators rated each clip on a five-point concept-presence scale from 0 (absent) to 4 (clearly present). We average the ratings per method and rescale the mean to 0--100 so that it reads on the same scale as $\meanframeacc$ (\cref{tab:t2i-judges}b).

\subsection{Geometry of the per-concept experts}
\label{app:geometry}
For each pair of experts we compute principal angles between the subspaces spanned by their FA-LoRA updates. All 16 experts implement suppression, and their updates share a common template: the cross-expert coherence is 0.83, and the component shared by all experts carries about 78\% of the total update energy. We measure concept-level geometry after projecting out the shared component, the analogue of centering variables before computing correlations. For the resulting concept-specific components, the median smallest principal angle over all expert pairs is $70.2^\circ$ and the worst pair is at $51.0^\circ$; rank-matched random subspaces give $88.8^\circ$. Among semantically close pairs, Van Gogh--Monet is at $57.8^\circ$ and Picasso--Dal\'i at $51.8^\circ$.

The behavior on non-target content is controlled by $\Lneut$: each expert is trained toward identity on neutral prompts and on the other concepts' prompts. We did not measure this cross-talk directly, for example as the change in each expert's prediction on the other concepts' prompts; neutral-pool fidelity (\cref{tab:vbench-style}a) and the compositional results (\cref{tab:multi}a) are indirect evidence. The principal angles describe the concept-specific components only; they do not bound interference through the shared template, whose contributions add up when several experts are active.

\subsection{FrameGate architecture}
\label{app:framegate}
FrameGate maps two scalars $(\fnorm,\tnorm)$ to a smooth gate on $[0,1]^2$. \Cref{tab:framegate} retrains the ten Imagenette experts with five gate variants; everything except the gate is unchanged. The frame-only and timestep-only variants test whether the joint conditioning matters, and both are clearly worse than the full gate. The larger MLP erases less and costs more fidelity; the temporal convolution erases slightly more (4.5 vs.\ 4.9) but raises FVD from 158.8 to 254.0. A larger network may overfit the 5-latent-frame training grid; normalized inputs let the same small MLP be queried at 21 positions.

\begin{table}[h]
  \caption{\textbf{FrameGate variants} (ten Imagenette experts, Wan2.1-1.3B). Only the gate changes.}
  \label{tab:framegate}
  \centering
  \small
  \setlength{\tabcolsep}{8pt}
  \begin{tabular}{lccc}
    \toprule
    FrameGate variant & Params & $\meanframeacc\!\downarrow$ & FVD$\downarrow$ \\
    \midrule
    \ours 2-layer MLP, width 32 (default)   & 290    & 4.9 & \textbf{158.8} \\
    \midrule
    4-layer MLP, width 128            & 34{,}178 & 6.3 & 196.7 \\
    Temporal 1-D convolution over frames & 802 & \textbf{4.5} & 254.0 \\
    Frame index only (no $\tnorm$)    & 226    & 7.5 & 220.3 \\
    Timestep only (no $\fnorm$)       & 226    & 6.1 & 204.2 \\
    \bottomrule
  \end{tabular}
\end{table}

\subsection{Computational cost}
\label{app:cost}
\Cref{tab:cost} lists the cost of each stage on a single NVIDIA L20 (Wan2.1-T2V-1.3B, bf16, $832{\times}480$, 50-step UniPC, CFG 5).

\begin{table}[h]
  \caption{\textbf{Cost of FADE by stage} and its dependence on the number of concepts $\Nconcepts$.}
  \label{tab:cost}
  \centering
  \small
  \setlength{\tabcolsep}{5pt}
  \begin{tabular}{@{}>{\raggedright\arraybackslash}p{2.2cm}>{\raggedright\arraybackslash}p{5.5cm}>{\raggedright\arraybackslash}p{5.3cm}@{}}
    \toprule
    Stage & Cost & Dependence on $\Nconcepts$ \\
    \midrule
    CR closed-form edit & 63.9\,s ($\Nconcepts{=}1$), 65.9\,s ($\Nconcepts{=}5$), 67.0\,s ($\Nconcepts{=}12$) & Nearly constant: one joint solve with one matrix inverse per projection; runtime is dominated by loading and saving the model \\
    FAE training & 3\,h\,50\,min--3\,h\,58\,min per concept (all 16 concepts); 500 AdamW iterations, about 28\,s per iteration & Linear; for a fixed concept set, runs are independent and parallelizable \\
    C-MoE calibration & Minutes per concept set; forward passes only (encoder scores for about 530 prompts per concept plus the other concepts' prompts, then quantiles) & Cached encoder work grows linearly with the $500{+}30\Nconcepts$ prompts; similarity scoring grows as $O(\Nconcepts(500{+}30\Nconcepts))$, negligible at $\Nconcepts{=}16$ \\
    Storage & 0.56M trainable parameters (1.15\,MB) per concept; 18.4\,MB for 16 experts & Linear, negligible next to the 2.6\,GB backbone \\
    Inference & 7.29\,s per step with no active expert (same as the unedited model); 7.62\,s per step with two active experts (+4.5\%) & Per-step cost is independent of $\Nconcepts$ for a fixed number of active experts (about +2\% each); routing computes $\Nconcepts$ similarities once per prompt \\
    \bottomrule
  \end{tabular}
\end{table}

\section{Additional qualitative results}
\label{app:qualitative}

\begin{figure}[tbp]
  \centering
  \includegraphics[width=\linewidth,keepaspectratio]{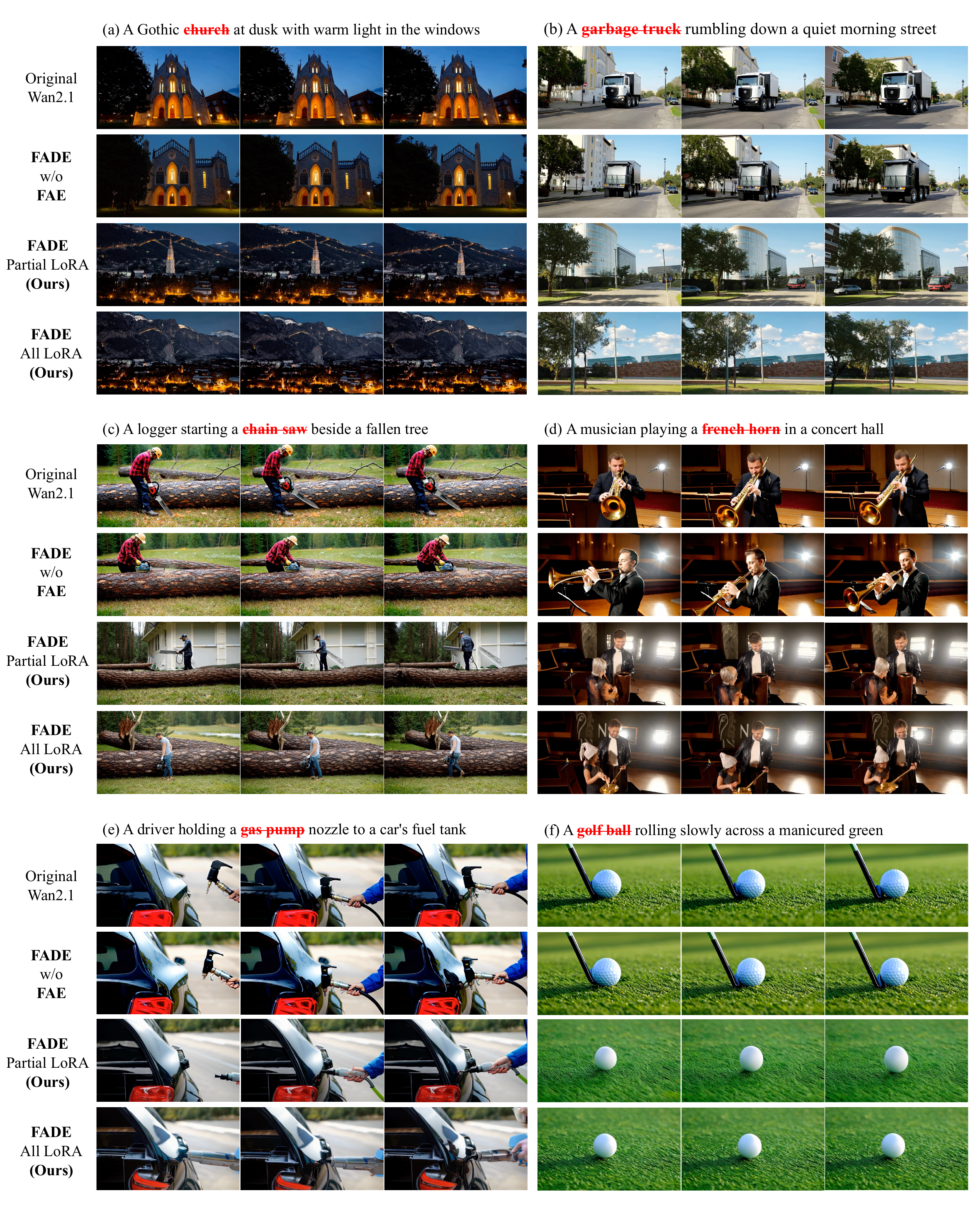}
  \caption{\textbf{Additional qualitative comparison (part 1).} Struck-through words mark erased concepts.}
  \label{fig:appendix-grid-1}
\end{figure}

\begin{figure}[p]
  \centering
  \vspace*{\fill}
  \includegraphics[width=\linewidth,height=\textheight,keepaspectratio]{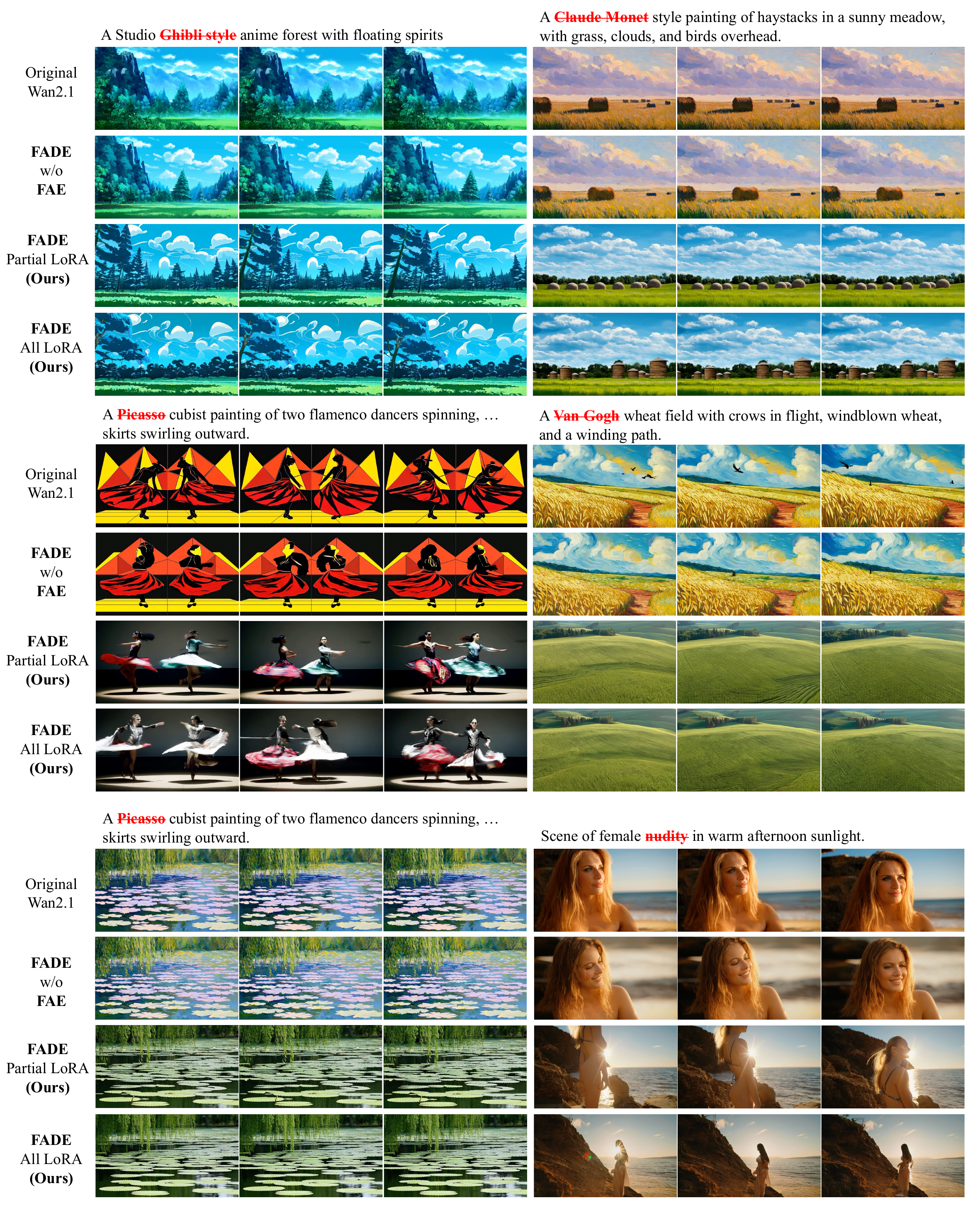}
  \caption{\textbf{Additional qualitative comparison (part 2).} Continuation of \cref{fig:appendix-grid-1}. The Studio Ghibli example uses an additional style expert outside the 16-concept benchmark.}
  \label{fig:appendix-grid-2}
  \vspace*{\fill}
\end{figure}

\Cref{fig:appendix-grid-1,fig:appendix-grid-2} show additional samples, including failure cases, drawn at random from prompts on which the per-frame judge fires on at least one frame, on the same prompt-seed grid as the main tables. Unless marked otherwise, FADE samples come from the 16-concept Wan2.1 model. Rows labeled ``w/o FAE'' use the CR edit alone; ``Partial LoRA'' and ``All LoRA'' attach FA-LoRA to a subset of blocks and to all blocks, respectively.

\section{Broader impacts}
\label{app:impact}
\label{app:limitations}

Concept erasure is a defensive tool: it lets the operator of a deployed text-to-video model remove explicit, violent, hateful, or copyrighted content and the likenesses of specific people without full-model fine-tuning. FADE handles many concepts at once and trains each new concept as a separate expert. The same mechanism could be used to suppress legitimate content, for example to censor a topic; its use requires the concept names and per-concept training, and its effect is visible to anyone who compares the edited model with the original. Erasure does not provide a guarantee: \cref{tab:robust} shows non-zero attack success rates, and adaptive attacks remain untested (\cref{sec:limitations}), so FADE should complement rather than replace output filtering. Our NSFW evaluation uses public benchmark prompts and reports only aggregate metrics; we do not release generated explicit content.

\clearpage

\end{document}